%% file: main.tex
\documentclass{article} 
\usepackage{iclr2026_conference,times}

\input{math_commands.tex}

\usepackage[pagebackref=true,colorlinks,citecolor=brown]{hyperref}
\renewcommand{\paragraph}[1]{\noindent\textbf{#1}}
\usepackage{caption}
\usepackage{url}
\usepackage{graphicx}
\usepackage{caption}
\usepackage{multirow}
\usepackage{booktabs}
\usepackage{xcolor}
\usepackage[dvipsnames]{xcolor} 
\usepackage{xspace}             
\usepackage{wrapfig}
\usepackage{algorithm}
\usepackage[noend]{algpseudocode} 
\usepackage{listings}
\usepackage{subcaption}
\usepackage{enumitem}
\usepackage{tikz}
\usetikzlibrary{shapes.geometric}
\usepackage[utf8]{inputenc}
\definecolor{xsienna}{rgb}{0.75,0.33,0.1}

\vspace{-20pt}
\title{Geometry-aware Policy Imitation}

\iclrfinalcopy
\author{Yiming Li\textsuperscript{$1,2$} 
\hspace{1.1em}
Nael Darwiche\textsuperscript{$2$}
\hspace{1.1em}
Amirreza Razmjoo\textsuperscript{$1,2$}
\hspace{1.1em}
Sichao Liu\textsuperscript{$2$}
\AND
Yilun Du\textsuperscript{$3$}
\hspace{1.1em}
Auke Ijspeert\textsuperscript{$2$}
\hspace{1.1em}
Sylvain Calinon\textsuperscript{$1,2$}
\\[1.5ex]
~$^1$Idiap Research Institute \ ~$^2$EPFL \  ~$^3$Harvard University
}

\begin{document}

\maketitle

\vspace{-15pt}
\begin{abstract}
We propose a Geometry-aware Policy Imitation (GPI) approach that rethinks imitation learning by treating demonstrations as geometric curves rather than collections of state–action samples. From these curves, GPI derives distance fields that give rise to two complementary control primitives: a progression flow that advances along expert trajectories and an attraction flow that corrects deviations. Their combination defines a controllable, non-parametric vector field that directly guides robot behavior. This formulation decouples metric learning from policy synthesis, enabling modular adaptation across low-dimensional robot states and high-dimensional perceptual inputs. GPI naturally supports multimodality by preserving distinct demonstrations as separate models and allows efficient composition of new demonstrations through simple additions to the distance field. We evaluate GPI in simulation and on real robots across diverse tasks. Experiments show that GPI achieves higher success rates than diffusion-based policies while running 20× faster, requiring less memory, and remaining robust to perturbations. These results establish GPI as an efficient, interpretable, and scalable alternative to generative approaches for robotic imitation learning. Project website: \href{https://yimingli1998.github.io/projects/GPI/}{https://yimingli1998.github.io/projects/GPI/}.
\end{abstract}

\input{intro}
\input{method}

\input{exp}

\input{rw}
\input{conclusion}



\newpage
\bibliography{iclr2026_conference}
\bibliographystyle{iclr2026_conference}

\input{appendix}

\end{document}

%% file: math_commands.tex
\usepackage{amsmath,amsfonts,bm}

\def\eqref#1{equation~\ref{#1}}

\def\1{\bm{1}}

\DeclareMathAlphabet{\mathsfit}{\encodingdefault}{\sfdefault}{m}{sl}
\SetMathAlphabet{\mathsfit}{bold}{\encodingdefault}{\sfdefault}{bx}{n}



%% file: intro.tex
\vspace{-5pt}
\section{Introduction}

Robots are increasingly expected to perform complex tasks in unstructured environments, ranging from dexterous manipulation to interactive collaboration. \emph{Imitation learning} offers a promising path toward this goal, as it enables robots to acquire policies directly from expert demonstrations without relying on explicit dynamics models or simulation. 
Existing imitation approaches can be grouped into three families. \emph{Explicit policies} treat imitation as supervised regression from states to actions~\citep{calinon2007learning}. They are fast at inference but struggle with multimodality and generalization. \emph{Implicit policies} learn energy functions over state--action pairs~\citep{florence2022implicit}, but are hard to train and slow to optimize at deployment. \emph{Generative policies}, such as diffusion or flow-matching models~\citep{chi2023diffusion,lipmanflow}, excel at modeling multimodality but remain computationally heavy and brittle under distribution shifts. Despite their differences, all three approaches compress demonstrations into parametric models that must be retrained to incorporate new data and that often discard the geometric structure underlying expert behavior.

\begin{figure}[t]
\vspace{-20pt}
    \centering
    \includegraphics[width=0.9\linewidth]{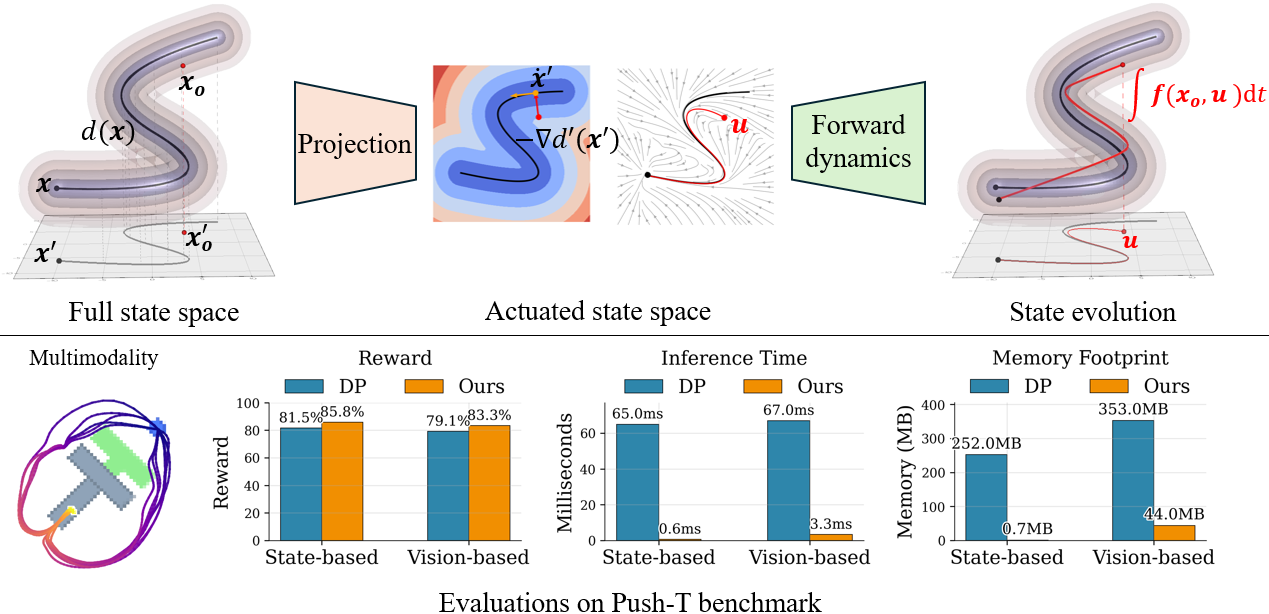}
    \vspace{-3pt}
\caption{
    \textbf{Overview of Geometry-Aware Policy Imitation (GPI).}  
    GPI treats demonstrations as geometric curves that induce distance fields in the full state space.  
    (\textbf{Top}) The state space is projected onto the robot’s actuated subspace, where control is applied. The projected distance field gives rise to two complementary flows: an \emph{attraction flow} from the negative gradient (red arrow) and a \emph{progression flow} from trajectory tangents (yellow arrow). Together, they define a dynamical system that reduces the distance to demonstrations and advances along them, thus imitating expert behavior. The resulting action $\bm{u}$ is executed through the system’s dynamics, yielding state evolution $\int f(x,u)\,dt$ in the full state space.     Multiple demonstrations can be composed naturally via Boolean operations on distance fields.   
    Despite unknown system dynamics, the resulting trajectory aligns closely with the most similar demonstration as determined by the distance metric. (\textbf{Bottom}) On the PushT benchmark, GPI achieves multimodal imitation with a higher reward, runs $20$--$100\times$ faster than diffusion policies (DDIM with 10 steps), and requires substantially less memory.  
}
    \label{fig:overview}
    \vspace{-20pt}
\end{figure}

We argue that imitation learning can be made more direct, interpretable, and efficient by adopting a \emph{geometric approach}. At its core, imitation means: (i) following the expert’s direction of motion, while (ii) approaching expert states as closely as possible. Viewed this way, a demonstration is not just a collection of samples but a \emph{geometric curve} in state space, annotated with tangents that indicate expert actions. This perspective motivates our approach, \textbf{Geometry-Aware Policy Imitation (GPI)}. GPI represents demonstrations as \emph{distance fields} that can be projected onto the robot’s actuated subspace, where control is applied. From these fields naturally emerge two complementary primitives: a \emph{progression flow} that advances along expert trajectories, and an \emph{attraction flow} that pulls current states toward them. Superimposing these flows defines a controllable vector field that drives imitation \citep{Li25MPDS}. This approach provides an approximation that reduces deviation while advancing along expert behaviors (Figure~\ref{fig:overview}). 
In addition, the policy is guided by a distance field composition that retrieves flow fields from the most similar demonstrations, promoting coherent behavior and enabling robustness even under unknown dynamics. 
 

A key strength of GPI is its \emph{decoupling} of imitation into two modular components:  
(i) \textbf{metric learning}, which defines how states are represented and compared; and  
(ii) \textbf{behavior synthesis}, which constructs policies directly from distance and flow fields.  
This separation offers substantial flexibility: low-dimensional states can use Euclidean or geodesic distances, while high-dimensional observations can rely on latent embeddings from pretrained or task-specific encoders. 
Policy synthesis itself is non-parametric and lightweight, enabling efficient composition of demonstrations without retraining and supporting multimodality by preserving distinct trajectories as separate flows~\citep{Pari-RSS-22}.  
Moreover, because GPI only requires a state representation that supports distance computation, rather than directly fitting a full policy function, the learning problem is considerably simpler than in generative models. 
Lightweight encoders are typically sufficient, which reduces training complexity and enables fast inference at deployment.  


We evaluate GPI extensively in both simulation and on real robots.  
In simulation, we benchmark across diverse domains---including planar pushing, 6-DoF manipulation, and dexterous hand control---with state spaces ranging from low-dimensional control vectors to raw vision inputs.  
For visual observations, we study multiple feature representations, from pretrained encoders to self-supervised embeddings.  
On real hardware, we demonstrate GPI on both a Franka arm and the Aloha bimanual system, showing that it scales robustly beyond controlled environments.  

In summary, our contributions are:  
\begin{itemize}[leftmargin=2em, nosep]
    \item[{\bf i)}] \textbf{Geometry-Aware Policy Imitation (GPI)}, which represents demonstrations as geometric curves that induce composable distance fields, providing a unified representation for both metric reasoning and action synthesis;  
    
    \item[{\bf ii)}] \textbf{A simple and modular formulation}, where state representation relies only on a suitable distance metric and action synthesis is realized through compositions of control primitives. 
    Both components are lightweight, flexible, and grounded in well-studied principles;  

    \item[{\bf iii)}] \textbf{Extensive validation} in simulation and on real robots, showing that GPI achieves higher performance and enables efficient policy imitation---over 20$\times$ faster than state-of-the-art diffusion policies---while remaining interpretable and multimodal. 

\end{itemize}

%% file: method.tex
\vspace{-5pt}
\section{Geometry-Aware Policy Imitation}

GPI constructs policies directly from demonstrations by representing them as geometric curves in state space. 
Each demonstration induces a distance field that encodes state similarity and gives rise to two complementary control primitives:  
(i) a \emph{progression flow} that advances along demonstrated motions, and  
(ii) an \emph{attraction flow} that corrects deviations by pulling states toward the trajectory.  
Their superposition defines a dynamical system that imitates expert behavior. 
Local policies derived from individual demonstrations are then composed via distance-based weighting, producing a coherent global policy that is efficient, interpretable, and robust to perturbations. 
Figure~\ref{fig:overview}-\textit{top} illustrates these components schematically.


\subsection{Method}
\label{sec:method}

We are given $N$ expert demonstrations 
$\mathcal{D} = \{\Gamma^{(i)}\}_{i=1}^N$, 
where each $\Gamma^{(i)}$ is a trajectory consisting of a sequence of states and actions
\begin{equation}
\Gamma^{(i)} = \{(\bm{x}_t^{(i)}, \bm{u}_t^{(i)})\}_{t=0}^{T_i},
\end{equation}
with states $\bm{x}_t^{(i)} \in \mathcal{X}$, actions $\bm{u}_t^{(i)} \in \mathcal{U}$, and horizon $T_i$.  

\paragraph{State and actuated subspace.}
A state $\bm{x}$ may include both environment variables (e.g., object poses, images) that are unactuated, and robot variables that are directly actuated by control inputs.  
We denote by $\bm{x}' = P(\bm{x})$ the projection of $\bm{x}$ onto the actuated subspace $\mathcal{X}' \subseteq \mathcal{X}$, where $P:\mathcal{X} \to \mathcal{X}'$ is the projection operator.  
Each trajectory $\Gamma^{(i)}$ can then be viewed as a geometric curve in state space, which induces a \emph{distance field} $d(\bm{x}_o \mid \Gamma^{(i)})$ measuring the proximity between a query state $\bm{x}_o$ and the demonstration.  

\paragraph{Action space.}
We assume \emph{velocity control} in the actuated subspace, i.e., $\bm{u}_t=\dot{\bm{x}}'_t$.
Each demonstration $\Gamma^{(i)}$ then defines a curve ${\bm{x}^{(i)}_t}$ whose actions ${\bm{u}^{(i)}_t}$ are the tangent directions in $\mathcal{X}'$.
Velocity control is used here for clarity, but it is not a prerequisite: the formulation extends naturally to accelerations or torques, which can be executed through the robot’s kinematics or dynamics models.  

\paragraph{Policy as flow field in actuated space.}  
From the distance field $d(\bm{x}_o \mid \Gamma^{(i)})$ induced by a demonstration $\Gamma^{(i)}$, we derive two complementary flows in the actuated subspace:  
\emph{Progression flow}, given by the demonstrated tangent action  
$\bm{u}^{(i)}_{\kappa(\bm{x}_o)} = \dot{\bm{x}}'^{(i)}_{\kappa(\bm{x}_o)}$,  
which advances along the expert trajectory; and  
\emph{Attraction flow}, obtained from the partial derivative of the distance field with respect to actuated coordinates,  
$-\nabla_{\bm{x}'_o} d(\bm{x}_o \mid \Gamma^{(i)})$,  
which corrects deviations by pulling states back toward demonstrations.  
Their superposition defines a policy in the actuated subspace:
\begin{equation}
\label{eq:ds}
\pi_i(\bm{x}_o) 
= \lambda_1(\bm{x}_o)\,\bm{u}^{(i)}_{\kappa(\bm{x}_o)} 
- \lambda_2(\bm{x}_o)\,\nabla_{\bm{x}'_o} d(\bm{x}_o \mid \Gamma^{(i)}),
\end{equation}
where $\kappa(\bm{x}_o) = \arg\min_t d(\bm{x}_o, \bm{x}^{(i)}_t)$ denotes the nearest demonstrated state,  
and $\lambda_1,\lambda_2 \geq 0$ are weights—either constant or distance-dependent chosen so that attraction dominates far from demonstrations, while progression dominates near them. This policy has been shown to yield a stable first-order dynamical system that asymptotically converges to the demonstrated trajectory if the state and action variables are continuous~\citep{Li25MPDS}\footnote{See Appendix~\ref{app:convergence} for the proof.}. This can be achieved by representing a discrete trajectory with continuous functions such as splines. Thus, the robot’s behavior remains robust, predictable, and safe even under environmental changes or perturbations.


\paragraph{Composition across demonstrations.}
To obtain a global policy, we compose local flow-based policies across multiple demonstrations.  
Given the $K$ nearest demonstrations, the global policy is
\begin{equation}
\pi(\bm{x}_o) = \sum_{i=1}^K w_i(\bm{x}_o)\,\pi_i(\bm{x}_o),
\qquad
w_i(\bm{x}_o) = \frac{\exp\!\big(-\beta\, d(\bm{x}_o \mid \Gamma^{(i)})\big)}
{\sum_{j=1}^K \exp\!\big(-\beta\, d(\bm{x}_o \mid \Gamma^{(j)})\big)},
\end{equation}
where $\pi_i(\bm{x}_o)$ is the local policy induced by demonstration $\Gamma^{(i)}$,  
$d(\bm{x}_o \mid \Gamma^{(i)})$ is the distance from query state $\bm{x}_o$ to the trajectory $\Gamma^{(i)}$, and $\beta > 0$ is a temperature parameter controlling the sharpness of selection.  
This distance-based composition ensures that flows are retrieved from the most relevant demonstrations, yielding coherent behavior even under unknown dynamics. A detailed description of GPI is provided in Algorithm~\ref{alg1} (Appendix~\ref{app:gpi_algorithm}).


\subsection{Choice of Distance Metric}

A central design choice in GPI is the distance metric $d(\bm{x}_o \mid \Gamma^{(i)})$, which measures the similarity between a query state and a demonstration. The state naturally consists of two complementary parts: the robot-actuated variables (e.g., joint angles, end-effector pose) and the environment-related variables (e.g., object poses, images). Accordingly, the distance metric can be decomposed into a robot feature $d_{\text{rob}}$ and an environment feature $d_{\text{env}}$, where the former also shapes the attraction flow in actuated space and the latter only influences demonstration selection and weighting.

\textbf{Robot distance $d_{\text{rob}}$.}  
For joint or end-effector positions $\bm{x} \in \mathbb{R}^n$, Euclidean distance is standard:
\begin{equation}
d_{\text{Euc}}(\bm{x}_1,\bm{x}_2) = \|\bm{x}_1 - \bm{x}_2\|_2.
\end{equation}  
For end-effector orientations represented as quaternions, geodesic distances on $S^3$ respect rotational geometry:
\begin{equation}
d_{\text{quat}}(\bm{x}_1,\bm{x}_2) = 2 \arccos\!\left(|\langle \bm{x}_1, \bm{x}_2 \rangle|\right).
\end{equation}  
These two cases cover the most common representations in joint space and task space for robotics.  

\begin{figure}[t]
    \centering
    \includegraphics[width=1.0\textwidth]{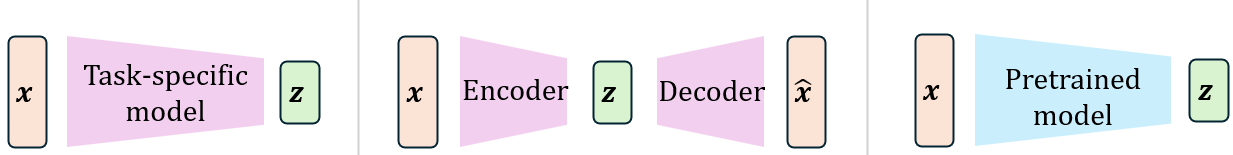}
    \caption{\textbf{Typical ways to obtain latent embedding $\bm{z}$ from raw inputs $\bm{x}$.} (i) train a task-specific lightweight model to capture task-relevant features; (ii) use a VAE to learn task-agnostic features; or (iii) apply a pretrained model to obtain features without additional training.}
    \label{fig:visual_representations}
    \vspace{-6mm}
\end{figure}

\textbf{Environment distance $d_{\text{env}}$.}  
This compares task-relevant but indirectly controllable variables, such as object poses or scene images.  
For low-dimensional object poses, $d_{\text{env}}$ can be computed with Euclidean or geodesic distances, reusing the formulations above.  
For high-dimensional observations, it is common to define $d_{\text{env}}$ in a latent space.
Let $\bm{z}=\Psi(\bm{x})$ denote the latent embedding of $\bm{x}$. Then
\begin{equation}
d_{\text{env}}(\bm{x}_1,\bm{x}_2)=d_{\text{env}}\!\left(\bm{z}_1,\,\bm{z}_2\right),
\end{equation}
where $\bm{z}_1=\Psi(\bm{x}_1)$ and $\bm{z}_2=\Psi(\bm{x}_2)$ are latent embeddings produced by a parametric model $\Psi$ that maps raw observations to a latent space, and $d(\cdot,\cdot)$ denotes a suitable distance (e.g., Euclidean or cosine).
This formulation supports multiple sources of embeddings: (i) task-specific models, where $\bm{z}$ could encode predicted object poses or desired robot actions learned via supervision; (ii) latent variables from variational autoencoders (VAEs) trained with self-supervised objectives~\citep{kingma2013auto}; and (iii) pretrained vision or multimodal encoders such as SAM~\citep{kirillov2023segment}, DINO~\citep{simeoni2025dinov3}, and CLIP~\citep{radford2021learning}, see Figure~\ref{fig:visual_representations} for an overview.
Classical dimensionality-reduction methods, such as principal component analysis (PCA), can also be used to obtain a compact latent feature~\citep{hotelling1933pca}.

While both $d_{\text{rob}}$ and $d_{\text{env}}$ contribute to the overall distance metric, 
their roles differ: $d_{\text{env}}$ influences only the similarity ranking across demonstrations, 
whereas $d_{\text{rob}}$ additionally shapes the attraction flow in the actuated subspace.  
This decomposition makes explicit how environmental features guide demonstration selection, 
while robot features govern the actual corrective control.  

\subsection{A 2D Example}  

To illustrate GPI, we consider a simplified 2D setting where the state consists only of actuated variables $\bm{x}'$.  
This abstraction is common in kinematic planning tasks, 
where environment dynamics are ignored.  
In this case, the distance field reduces to the robot-related term, $d(\bm{x}_o) = d_{\text{rob}}(\bm{x}'_o)$,  
so that state evolution and policy flows are fully contained in the same space. While prior work typically trains diffusion or flow-matching models for policy generation in this setting \citep{jiang2025streaming}, GPI instead addresses the problem in a fully non-parametric manner, relying directly on the distance and flow fields.

Figure~\ref{fig:2d_example}(a) shows two demonstrations forming a Y-shaped pattern:
$\Gamma^{(1)}$ (green) and $\Gamma^{(2)}$ (blue) overlap initially and then diverge into separate branches.
Temporal progression is indicated by transparency from $t=0$ to $t=1$.
Each demonstration induces a Euclidean distance field whose valleys align with its trajectory; composing them yields a global
distance field (Figure~\ref{fig:2d_example}b) visualized as an energy landscape with dense corridors along the demos and a natural
decision boundary at the bifurcation. Figures~\ref{fig:2d_example}(c,d) show the resulting flow fields: each row includes the
single-demo flow (left) and the composed flow (both demos), with rollout trajectories overlaid on the energy landscape (right).
Panel (c) depicts the progression flow, which follows the local tangent of the nearest demonstration; Panel (d) augments this with
an attraction term that pulls states toward the trajectories, ensuring stable convergence. The rollout trajectories (red) show the integrated trajectories in two cases. From this perspective, diffusion policies perform well because their denoising steps implicitly induce an attraction flow toward demonstrations rather than relying solely on progression.

\begin{table*}[t] 
\centering 
\setlength{\tabcolsep}{2pt} 
\renewcommand{\arraystretch}{0.9}
\begin{tabular}{c@{\hspace{12pt}}ccc}
\includegraphics[width=0.23\linewidth]{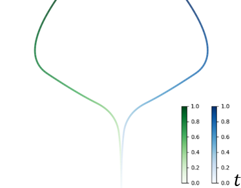} & 
\includegraphics[width=0.23\linewidth]{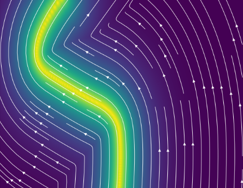} & 
\includegraphics[width=0.23\linewidth]{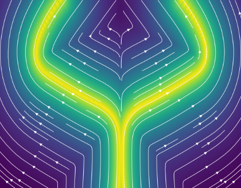} & 
\includegraphics[width=0.23\linewidth]{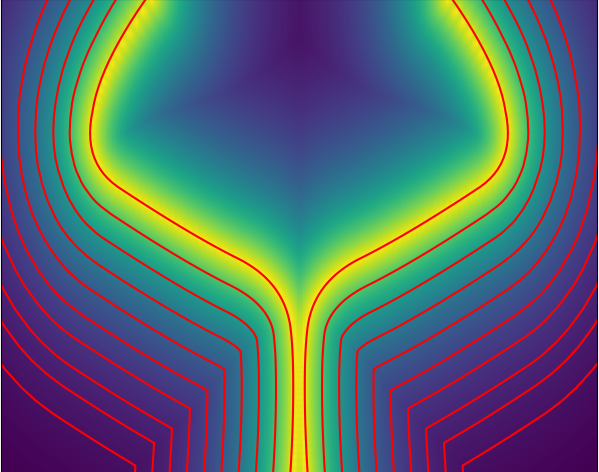} \\
(a) Demonstrations & 
\multicolumn{3}{c}{(c) Flow field with action $\bm{u}=\dot{\bm{x}}$} \\
\includegraphics[width=0.23\linewidth]{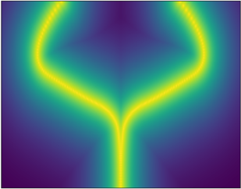} & 
\includegraphics[width=0.23\linewidth]{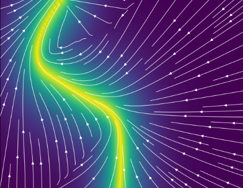} & 
\includegraphics[width=0.23\linewidth]{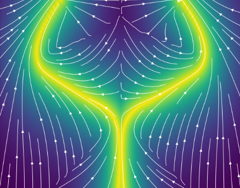} & 
\includegraphics[width=0.23\linewidth]{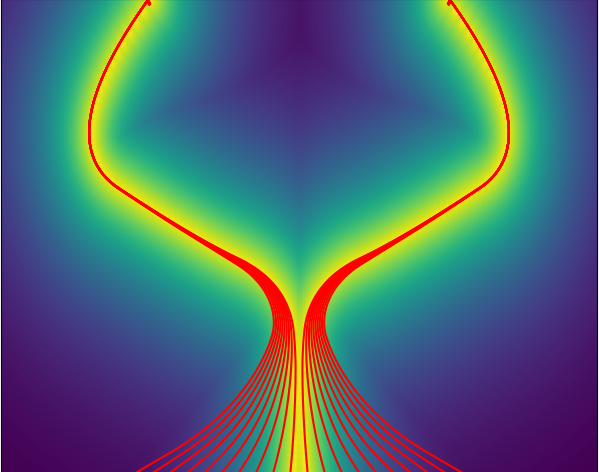} \\
(b) Energy landscape & 
\multicolumn{3}{c}{(d) Flow field with $\bm{u} = \lambda_1\dot{\bm{x}} - \lambda_2 \nabla_{\bm{x}} d$} \\
\end{tabular} 
\captionof{figure}{\textbf{From demonstrations to policy flows.}
(a) Demonstrations. (b) Energy from composed distances.
(c) Progression-only flow $\bm{u}=\dot{\bm{x}}$ may drift off the demonstrations.
(d) Adding attraction $\bm{u}=\lambda_{1}\dot{\bm{x}}-\lambda_{2}\nabla_{\bm{x}} d$ pulls states toward the demonstrations and along them, ensuring convergence.}

\label{fig:2d_example} 
\vspace{-6mm}

\end{table*}
\noindent By representing demonstrations as distance and flow fields, policy imitation shifts from fitting a parametric model to geometric reasoning grounded in similarity, curvature, and composition, yielding several benefits: \textbf{Efficiency}—new demonstrations enrich the distance field by adding basins of attraction without retraining, and inference reduces to distance evaluations plus weighted averaging of expert actions, making it lightweight and parallelizable; \textbf{Flexibility}—decoupling similarity measurement from action synthesis keeps the framework modular, allowing task-specific distance metrics and flow compositions; \textbf{Multimodality}—each demonstration defines its own distance and flow field, preserving distinct behaviors so the policy branches smoothly toward the nearest demonstrated mode instead of averaging conflicting actions; \textbf{Interpretability}—the distance metric reveals which demonstrations influence the current action, while actions remain a linear superposition of demonstrated behaviors and corrective flows, ensuring safe, bounded outputs.

%% file: exp.tex
\vspace{-5pt}
\section{Experimental Results}
\vspace{-3pt}

\subsection{Simulation Experiments}
We first evaluate GPI on the PushT benchmark, a widely adopted task in which a robot must push a T-shaped object into a target configuration~\citep{chi2023diffusion}. This environment is particularly suitable for evaluation: it has well-established baselines for comparison, requires handling inherently multimodal pushing strategies, and involves contact-rich dynamics that cannot be solved by simple kinematic planning.  

For state-based inputs, demonstrations consist of the agent position, the object position, and the object orientation. Distances are computed as a weighted combination of these components. The actuated subspace corresponds to the agent position, with its first-order derivative (velocity) serving as the action. Note that the original environment specifies actions in position control, which we adapt to velocity control for consistency with our flow-based formulation. Control policies are synthesized from the flow fields induced in the actuated subspace by corresponding demonstrations, and then executed in the environment with unknown interaction dynamics. For vision-based inputs, the state comprises the agent pose and an RGB image. Distances are computed jointly over the agent pose and an image embedding. To align with the state-based formulation, we train a lightweight task-specific model to produce the image embedding as the predicted object pose.


\begin{table}[h]
\centering
\vspace{-5pt}
\caption{Performance comparison on Push-T (state-based vs. vision-based). }
\label{tab:push_t}
\resizebox{\linewidth}{!}{
\begin{tabular}{l
                c
                c
                c
                c
                c
                c}
\toprule
& \multicolumn{3}{c|}{Push-T (state-based)} 
& \multicolumn{3}{c}{Push-T (vision-based)} \\
\cmidrule(lr){2-4} \cmidrule(lr){5-7}
Method 
& (Avg./Max.) score
& Training / Inference Time 
& Memory 
& (Avg./Max.) score
& Training / Inference Time  
& Memory \\
\midrule
DDPM  
& 82.3 / 86.3 & 1.0\,h / 641\,ms & 252\,MB
& 80.9 / 85.5 & 2.5\,h / 647\,ms & 353\,MB \\
DDIM
& 81.5 / 85.1 & 1.0\,h / 65\,ms  & 252\,MB
& 79.1 / 83.1 & 2.5\,h / 67\,ms  & 353\,MB \\
GPI (Ours)             
& \textbf{85.8 / 89.0} & \textbf{0\,h / 0.6\,ms} & \textbf{0.7\,MB}
& \textbf{83.3 / 86.9} & \textbf{0.3\,h / 3.3\,ms} & \textbf{44\,MB} \\
\bottomrule
\end{tabular}
}
\vspace{-5pt}
\end{table}
Experiments were conducted on an NVIDIA RTX 3090 GPU. Further details appear in Appendices~\ref{app:pushT state} and~\ref{app:pushT vision}.
 We report performance using three complementary metrics: (i) \textit{Average / maximum reward}, evaluated over multiple random seeds and environment variations, following the same protocol as the baselines; (ii) \textit{time}, including training time and per-step inference time; and (iii) \textit{memory footprint}, including memory cost for model parameters and stored demonstrations. Results are summarized in Table~\ref{tab:push_t}. We compare GPI with Diffusion Policy~\citep{chi2023diffusion} using both 100-step Denoising Diffusion Probabilistic Models (DDPM)~\citep{ho2020denoising} and 10-step Denoising Diffusion Implicit Models (DDIM)~\citep{song2020denoising}. Note that, unlike diffusion policies which require predicting an action horizon (e.g., $H=8$), our approach naturally supports reactive planning and operates with horizon $H=1$. GPI achieves higher success rates than the diffusion policy while being substantially more efficient.

In the state-based setting, inference involves only low-dimensional, non-parametric distance evaluations and flow field composition, resulting in a latency of $0.6\,\text{ms}$—nearly $100\times$ faster than Diffusion Policy with 10 DDIM denoising steps. Although GPI requires storing all demonstrations for distance measurement, the overall memory footprint remains lower than that of training large neural policies\footnote{See Appendix~\ref{app:memory_cost} for a detailed explanation.} Moreover, the underlying computations are lightweight and naturally parallelizable, further contributing to its efficiency. For vision-based inputs, we employ a ResNet-18 encoder trained solely for feature extraction rather than precise action prediction, which simplifies training and improves efficiency. As a result, training completes in only $0.3$ hours (compared to $2.5$ hours for Diffusion Policy) and inference runs at $3.3\,\text{ms}$ per step (compared to $67\,\text{ms}$ for Diffusion Policy). Memory requirements are also reduced, since we store only the lightweight encoder and latent embeddings of demonstrations rather than raw images or large policy networks. Additionally, this modular structure allows the visual encoder to be reused across different tasks.

\begin{wrapfigure}{r}{0.4\textwidth} %
    \centering
    \vspace{-20pt}
    \includegraphics[width=0.4\textwidth]{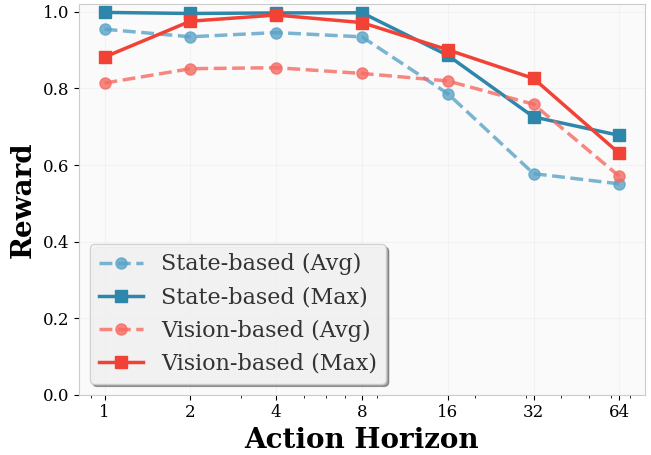}
    \vspace{-10pt}
    \caption{\textbf{Robustness to action horizons.}}
    \label{fig:action_horizon}
    \vspace{-20pt}
\end{wrapfigure}

We further conduct a series of ablations to highlight the distinctive properties of GPI:  

\textbf{Robustness.}  
We evaluate GPI’s robustness along three complementary dimensions.  

\textit{Planning horizon:} GPI is reactive by default ($H=1$), but it can also be extended to a receding-horizon scheme by updating the distance every $H$ steps. As shown in Figure~\ref{fig:action_horizon}, performance remains stable for horizons up to 16, showing GPI can operate either as a purely reactive controller (robust to external disturbances) or as a receding-horizon planner (with improved temporal consistency).  

\begin{figure}[h]
    \centering
    \includegraphics[width=0.8\linewidth]{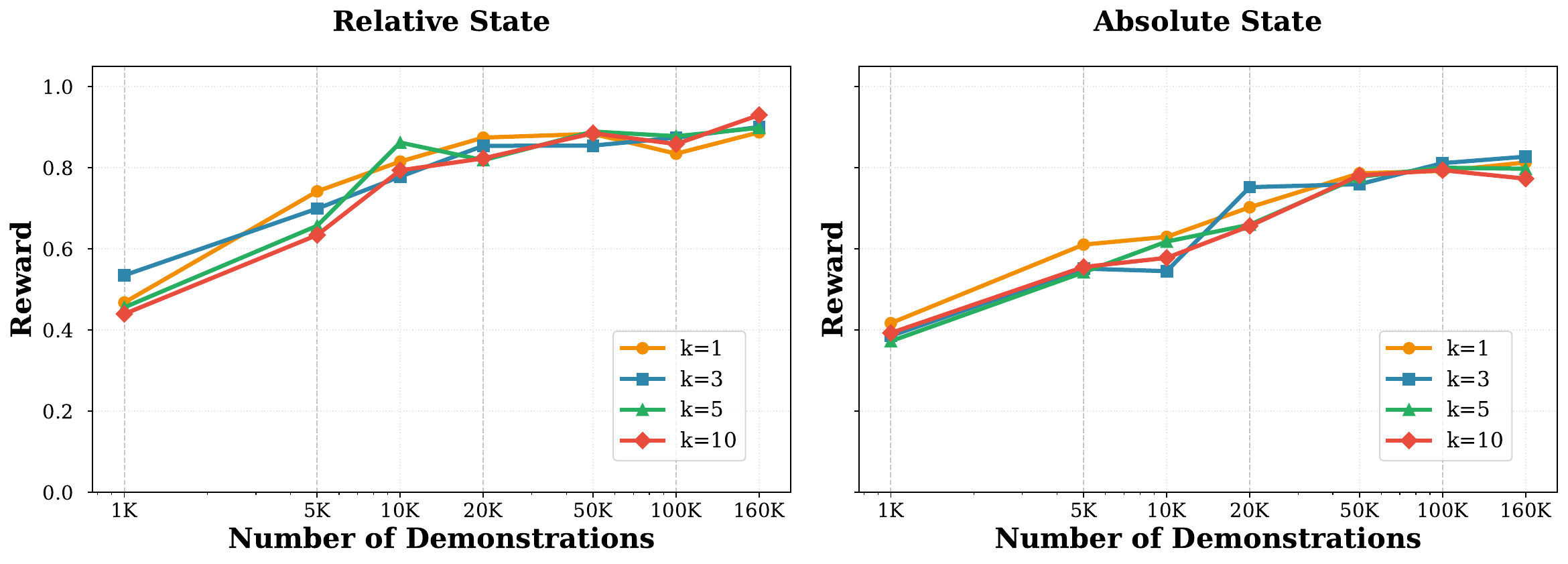}
\caption{\textbf{Robustness of GPI with respect to demonstrations, $K$ (neighbors), and state representations.}}
    \label{fig:action_subset_size}
    \vspace{-8mm}
\end{figure}

\textit{Number of neighbors:} In action composition, we compare $K=1,3,5,10$. As shown in Figure~\ref{fig:action_subset_size}, the curves are nearly overlapping in both relative and absolute state settings, confirming that performance is largely insensitive to the choice of $K$. This highlights the reliability of GPI’s local composition mechanism.  

\textit{State representation:} We compare object-centric (relative) and global (absolute) state formulations (Figure~\ref{fig:action_subset_size}). Both achieve strong performance, but relative states consistently yield slightly higher scores, especially in data-scarce regimes. This suggests that GPI is robust to representation choices, with relative states offering an advantage when demonstrations are limited.  
 
\textbf{Scalability with data sizes.}  A distinctive advantage of GPI is that, being non-parametric and training-free in the state-based setting, it enables direct study of how performance scales with the number of demonstrations, without the need for retraining. To this end, we augment the dataset with up to 160K samples regenerated from the original diffusion policy work and evaluate how performance evolves as the demonstration set grows. This setting is particularly suitable for GPI, since demonstration density directly influences both the distance query and the selection of actions in the composed policy. As shown in Figure~\ref{fig:action_subset_size}, success rates increase consistently as the dataset expands from 1K to 20K demonstrations, after which performance begins to saturate. This trend reveals two key insights: (i) larger demonstration sets provide denser coverage of the state space, thereby reducing approximation errors introduced by the chosen distance metric, and (ii) our approach can serve as a practical diagnostic tool—indicating how many demonstrations are sufficient to achieve reliable policy performance before training parametric models. The method also accommodates incremental incorporation of new demonstrations, without the need for full retraining.

\begin{wrapfigure}{r}{0.4\textwidth} %
    \centering
    \vspace{-20pt}
    \includegraphics[width=0.4\textwidth]{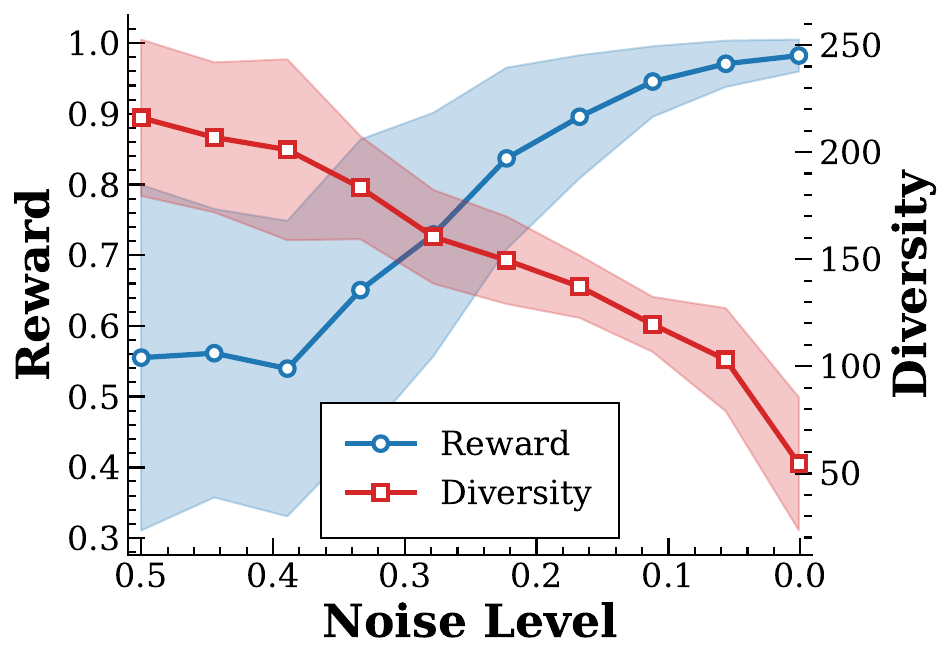}
    \vspace{-12pt}
    \caption{\textbf{Noise-level ablations for score and diversity.}}
    \label{fig:noise_comparison_plots}
    \vspace{-12pt}
\end{wrapfigure}

\textbf{Stochasticity and multimodality.}  
To induce stochasticity and multimodality, we inject Gaussian noise $\mathcal{N}(0,\sigma^2)$ into the query state in the actuated space (corresponding to the agent’s position). This perturbation alters the effective distance fields used in composition, thereby modifying the synthesized flow field and inducing multimodal behavior. In Figure~\ref{fig:noise_comparison_plots}, we compare the average score achieved under different noise levels. To quantify diversity, we measure the average distance among trajectories generated with different random perturbations sampled from the same noise distribution. The results show that larger noise values increase trajectory diversity but degrade performance, whereas smaller noise levels yield more deterministic behavior. Importantly, GPI exhibits multimodal behavior even under low noise (e.g., $\sigma=0.2$), as illustrated in Figure~\ref{fig:overview} (bottom left). Beyond Gaussian perturbations, stochasticity can also be enhanced by randomly subsampling the set of demonstrations at each inference time. We found that this strategy can improve performance in practice, for instance, by helping the robot escape from regions where it would otherwise become stuck.

\begin{wrapfigure}{r}{0.4\textwidth} %
    \centering
    \vspace{-20pt}
    \includegraphics[width=0.4\textwidth]{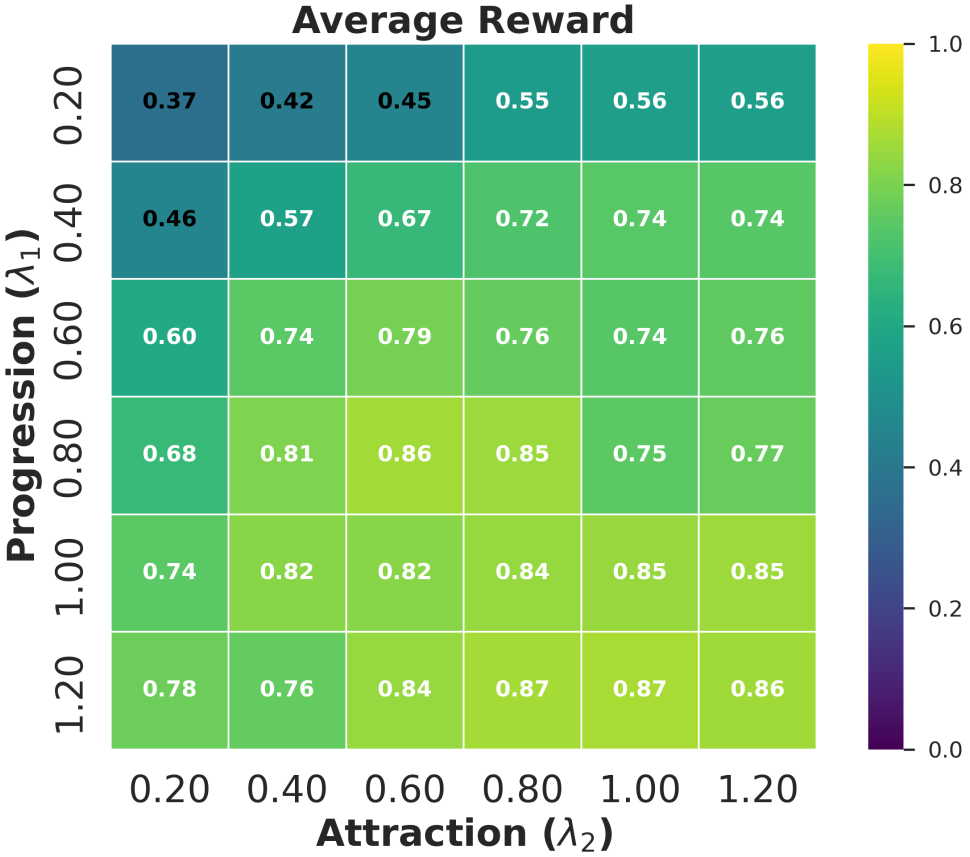}
    \vspace{-20pt}
    \caption{\textbf{Ablations on two control primitives.}}
    \label{fig:lambda_reward}
    \vspace{-20pt}
\end{wrapfigure}

\textbf{Natural composition of control primitives.} 
We interpret progression and attraction as two basic control primitives that can be naturally combined within the flow field. By varying their relative weights $(\lambda_1, \lambda_2)$, we interpolate between velocity-like (progression-driven) and position-like (attraction-driven) control. As shown in Figure~\ref{fig:lambda_reward}, GPI maintains consistently high scores across a wide range of weightings, demonstrating flexibility in composing these primitives at test time rather than relying solely on fixed neural network outputs. In this view, progression promotes forward motion and task advancement, while attraction provides goal alignment and stability.  

\textbf{Generalization across tasks.}  We evaluate GPI on RoboMimic (Lift, Can, Square)~\citep{robomimic2021} and Adroit (Door, Pen, Hammer, Relocate) benchmarks~\citep{rajeswaran2018learning}, spanning state spaces of 9–46 dimensions and action spaces of 7–30. GPI consistently matches or exceeds the performance of Diffusion Policy without requiring any parametric training (Table~\ref{tab:benchmarks}), demonstrating robust generalization across diverse domains. The snapshots of those tasks are shown in Figures \ref{fig:robomimic_exp} and \ref{fig:adroit_hand_task} (in Appendix \ref{app:other_tasks}) respectively. Additionally, we test GPI on 2D Maze task~\citep{chen2025diffusion, janner2022diffuser} and visualization results in shown in Figure~\ref{fig:2d_maze_res} in Appendix~\ref{app:2d_maze}.
\begin{table}[h]
\centering
\caption{Task description and performance on Robomimic and Adroit Hand benchmarks.}
\label{tab:benchmarks}
\resizebox{0.8\linewidth}{!}{
\begin{tabular}{llccc|cccc}
\toprule
 & & \multicolumn{3}{c|}{Robomimic} & \multicolumn{4}{c}{Adroit Hand} \\
\cmidrule(lr){3-5} \cmidrule(lr){6-9}
 & Task / Method & Lift & Can & Square & Door & Pen & Hammer & Relocate \\
\midrule
\multirow{3}{*}{Description} 
 & State Dim       & 9  & 16 & 16 & 39 & 45 & 46 & 39 \\
 & Action Dim      & 7  & 7  & 7  & 28 & 24 & 26 & 30 \\
 & Demonstrations  & 300 & 300 & 300 & 5000 & 5000 & 5000 & 5000 \\
\midrule
\multirow{3}{*}{Results}
 & DP & \textbf{1.00} & 0.94 & \textbf{0.87} & \textbf{1.00} & 0.89 & 0.83 & \textbf{0.91} \\
 & Ours             & \textbf{1.00} & \textbf{0.96} & 0.82 & \textbf{1.00} & \textbf{0.95} & \textbf{0.88} & \textbf{0.91 }\\
\bottomrule
\end{tabular}
}
\vspace{-10pt}
\end{table}

\paragraph{Generalization across visual representations.}
As discussed in Section~\ref{sec:method}, GPI naturally accommodates multiple choices of latent embeddings, including task-specific encoders, VAEs, and pretrained models.
We evaluate three variants on \textsc{PushT}:
(i) a ResNet feature~\citep{he2016deep} pretrained within the Diffusion Policy implementation, with PCA applied for dimensionality reduction;
(ii) an unsupervised variational autoencoder (VAE) trained solely on RGB images, serving as a task-agnostic feature extractor; and
(iii) a pretrained Segment Anything (SAM) model~\citep{kirillov2023segment} followed by a pose-estimation module whose predicted object pose serves as the embedding.
Implementation details are provided in Appendices~\ref{app:resnet_pca} (ResNet+PCA),~\ref{app:vae} (VAE) and~\ref{app:sam} (SAM).


\begin{wraptable}{r}{0.4\linewidth} 
\vspace{-\intextsep}                 
\centering
\caption{Performance of various visual representations on the pushT task.}
\label{tab:feature-extractors}
\begin{tabular}{@{}lc@{}}  
\toprule
\textbf{Feature Extractor} & \textbf{Avg. Score} \\
\midrule
Diffusion Policy   & 85 \\
Task-specific Head & 87 \\
\hline
ResNet+PCA         & 84 \\
VAE                & \textbf{88} \\
Pretrained SAM     & 41 \\
\bottomrule
\end{tabular}
\end{wraptable}

Results in Table~\ref{tab:feature-extractors} show that GPI with the same ResNet features followed by PCA achieves performance comparable to Diffusion Policy, which uses the same ResNet features with a diffusion head. 
Interestingly, a lightweight VAE encoder trained only for reconstruction also yields strong performance. 
A plausible explanation is that the KL regularizer encourages latents to stay near the prior \( \mathcal{N}(0,I) \), yielding a smoother latent space where linear interpolations tend to remain on-manifold. Notably, this VAE trains in \(\sim\!0.3\) hours and runs at \(\sim\!4\) ms per inference—similar to our task-specific head (Table~\ref{tab:push_t}). This highlights GPI’s robustness across vision features for non-parametric policy composition. In contrast, off-the-shelf SAM underperforms, likely due to sensitivity to segmentation quality and the downstream pose estimation module; we expect fine-tuning to improve results.

\begin{wrapfigure}{r}{0.62\textwidth} %
    \centering
    \vspace{-30pt}
    \includegraphics[width=0.62\textwidth]{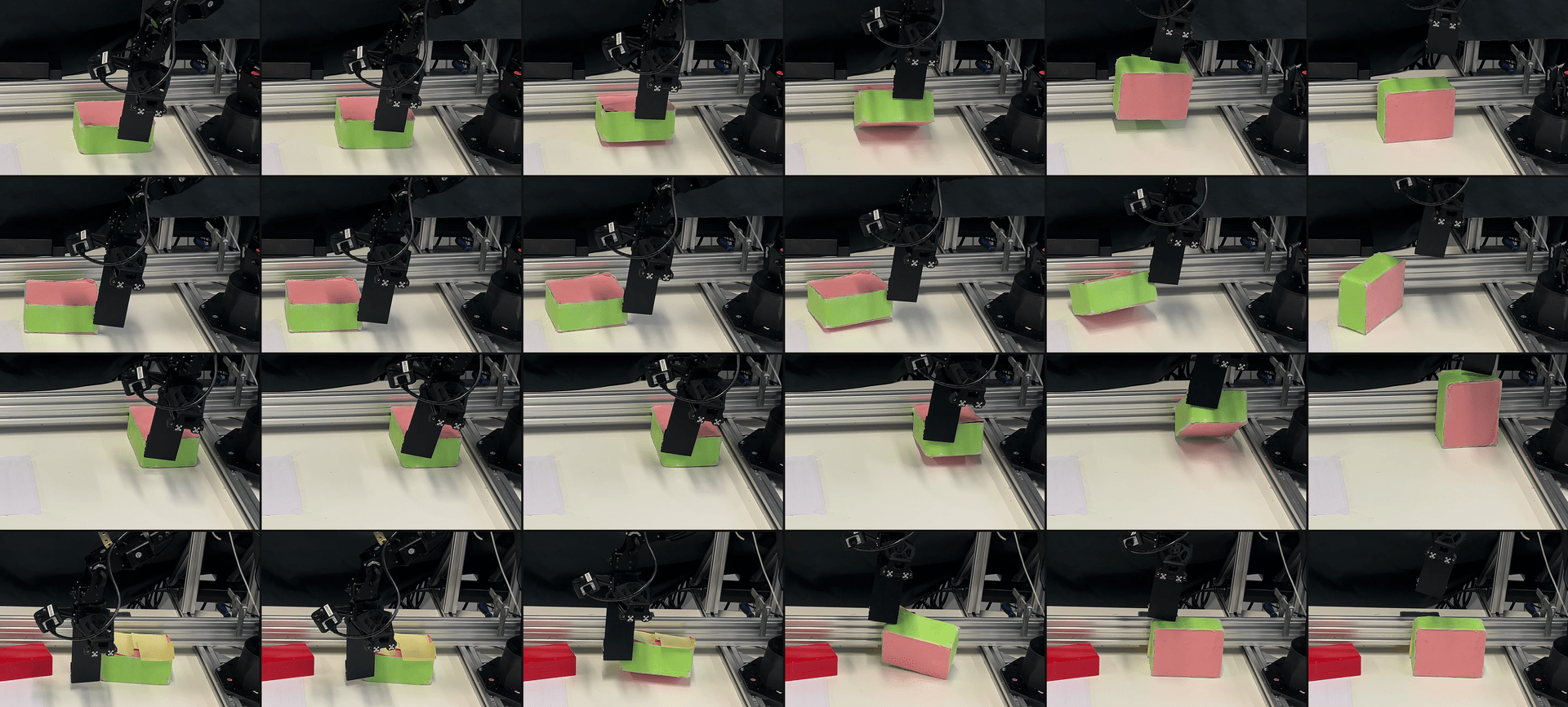}
    \vspace{-10pt}
    \caption{\textbf{Real-robot flipping task.} GPI successfully completes the task via multimodal behavior (Top 3 rows) and demonstrates robustness to visual disturbances (Bottom).
}
    \label{fig:robot_flip}
    \vspace{-15pt}
\end{wrapfigure}

\vspace{-5pt}
\subsection{Robot Experiments}
\vspace{-3pt}
To further evaluate GPI, we conduct robot experiments on two challenging tasks:

\textbf{(i) Box flip.}
The robot must flip a box by exploiting contacts among the end-effector, the box, and an aluminum crossbeam, which is challenging due to unknown, highly nonlinear dynamics. We collect $121$ demonstrations on an ALOHA platform~\citep{aldaco2024aloha}. The dataset contains over $50,000$ RGB images and action pairs. A lightweight neural network takes a raw RGB image as input and predicts an action; this predicted action serves as the image embedding. Distances are computed jointly over the robot joint configuration and the action embedding to construct the distance field, from which the flow field is derived for the robot's execution. We observe an inference time of approximately 7 ms and a memory footprint of 140 MB, comprising 139 MB for the feature-extraction model and 1 MB for storing latent features.

\textbf{(ii) Human–robot fruit handover.}
A human hands fruit to the robot. The robot must execute a smooth, anticipatory interaction while synchronizing its timing with the human and remaining robust to unpredictable motions and sensing noise. This task is run on a Franka robot.
\begin{wrapfigure}{r}{0.4\textwidth} %
    \centering
    \vspace{10pt}
    \includegraphics[width=0.4\textwidth]{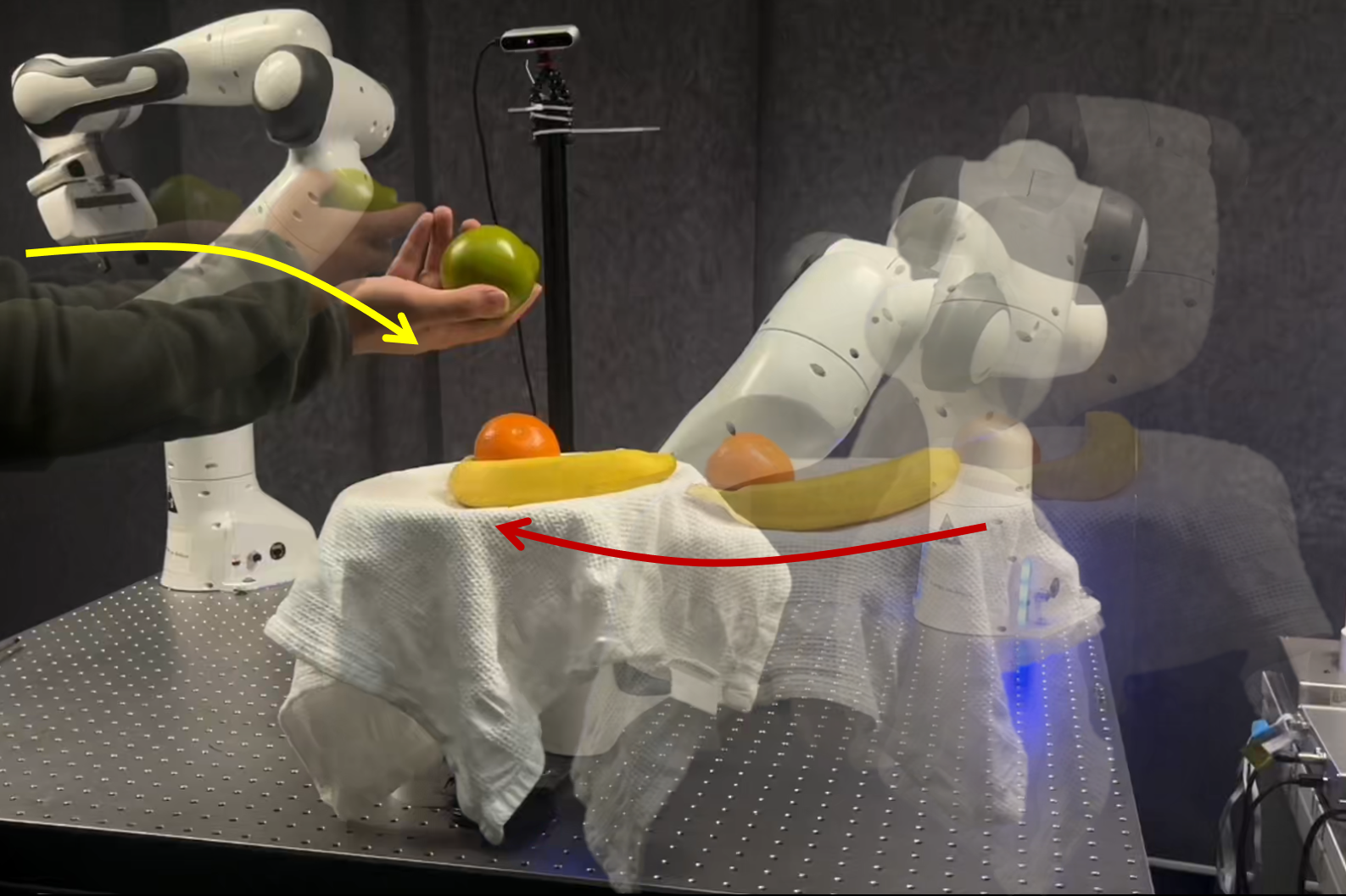}
    \vspace{-10pt}
    \caption{Real robot experiment on human-robot interaction task.}
    \label{fig:hri}
    \vspace{-20pt}
\end{wrapfigure}
We collect a single demonstration to align the robot’s motion phase with the human hand 
trajectory. At execution time, a pretrained CLIP model \citep{radford2021learning} provides a fruit-detection score, which we combine with the deviation from the demonstrated hand trajectory to define the distance field. This field determines the robot’s phase and progression; the robot follows the progression flow until the desired phase is reached, yielding synchronized and fluid handovers.

More details about the robot platform, experimental setup, and training details are illustrated in Appendices \ref{app:robot-flip} and \ref{app:hri}, respectively.
The robot behavior during two tasks is shown in Figures \ref{fig:robot_flip}, \ref{fig:hri} and the attached video. 

%% file: rw.tex
\vspace{-5pt}
\section{Related Work}
\vspace{-3pt}
\looseness=-1
Among approaches to acquiring robotic skills—reinforcement learning~\citep{sutton1998reinforcement} and optimal control~\citep{bertsekas1995dynamic}, imitation learning (IL)~\citep{osa2018algorithmic} stands out for not requiring explicit task models or cost functions, making it especially appealing when dynamics are hard to model. Even when such models exist, demonstrations can accelerate and improve solutions~\citep{Nair2017OvercomingEI,razmjoo2021optimal}.
Early approaches focus on time-dependent dynamical movement primitives, such as Dynamic movement primitives (DMP)~\citep{ijspeert2013dynamical} and Probabilistic Movement Primitives (ProMP)~\citep{paraschos2013probabilistic}, or time-independent dynamical systems~\citep{khansari2011learning}. They provide well-established approaches and efficient frameworks, but are usually limited in capturing complex, multi-modal demonstration patterns. Recent learning-based approaches, such as Implicit Behavior cloning and Diffusion policy, address this issue and have demonstrated impressive performance across a range of tasks~\citep{florence2022implicit,chi2023diffusion,zhang2024affordance}. However, these methods introduce challenges such as hard to train, slow inference, and need multi-step inference~\citep{lecun2006tutorial,du2019implicit,song2019generative,nijkamp2020ebm,zhang2024affordance}. GPI bridges dynamical systems and modern learning by representing demonstrations as distance fields—linking naturally to metric learning for high-level scene representations while inducing flow fields for low-level control. The closest prior, VINN~\citep{Pari-RSS-22}, learns visual representations via self-supervision and retrieves policies with $k$NN, achieving strong visual imitation. In contrast, GPI supports diverse latent representations and synthesizes policy flows—demonstrating effectiveness on tasks with complex dynamics.

%% file: conclusion.tex
\vspace{-5pt}
\section{Limitation and Conclusion}
\vspace{-3pt}

We present Geometry-aware Policy Imitation (GPI), which treats demonstrations as geometric curves that induce a distance field and policy flows. This perspective yields a simple, flexible, efficient, multimodal, and interpretable policy that composes behaviors and integrates with diverse latent representations. Our approach has a few limitations that are worth exploring in future work:

\paragraph{Choice of distance metric.} The metric is a primary design lever that shapes induced flows. Making it learnable and co-optimized with policy synthesis—optionally conditioned on task or context—can improve robustness and out-of-distribution generalization. Leveraging large models to provide task-relevant robotic features is especially promising~\citep{intelligence2025pi,barreiros2025careful}.

\paragraph{Scene dynamics and stability.} Our current results treat environment dynamics as unknown. A natural extension is to incorporate known or learned dynamics models and analyze when the resulting closed loop is stable and robust, e.g., via Lyapunov or contraction certificates with perturbation and model-mismatch bounds.

\paragraph{Scalability of demonstrations.} Although GPI stores only latent features, memory still scales linearly with the number of demonstrations. Future work could improve data efficiency with compact implicit distance parameterizations, while preserving geometric fidelity and fast retrieval.

%% file: appendix.tex
\clearpage

\appendix
\section*{Appendix}

\section{Convergence of the Flow Policy}
\label{app:convergence}

We prove convergence of the policy introduced in Section~\ref{sec:method}, which combines progression and attraction flows to form a stable dynamical system in the actuated subspace. For clarity, we rewrite the flow policy (\eqref{eq:ds}) as
\begin{equation}
\dot{\bm{x}} = \lambda_1 \dot{\bm{x}}_{t^\ast} - \lambda_2 \nabla d(\bm{x}),
\end{equation}
where $d(\bm{x})$ is the distance to the demonstration, $\nabla d(\bm{x})$ its gradient, $\dot{\bm{x}}_{t^\ast}$ the tangent velocity at the projection point $\bm{x}_{t^\ast}$, and $\lambda_1,\lambda_2 \geq 0$ weight progression and attraction.

We analyze stability using the Lyapunov function
\begin{equation}
V(\bm{x}) = \tfrac{1}{2} d^2(\bm{x}) \geq 0,
\end{equation}
which vanishes only on the demonstration. Its time derivative is
\begin{equation}
\dot{V}(\bm{x}) = d(\bm{x}) \, \nabla d(\bm{x})^\top \dot{\bm{x}}.
\end{equation}
Substituting the dynamics gives
\begin{equation}
\dot{V}(\bm{x}) = d(\bm{x}) \, \nabla d(\bm{x})^\top 
\big( \lambda_1 \dot{\bm{x}}_{t^{*}} - \lambda_2 \nabla d(\bm{x}) \big).
\end{equation}

To simplify this expression, we use the fact that the projection point $\bm{x}_{t^\ast}$ is defined as the minimizer of the squared distance
\begin{equation}
\|\bm{x}_{t} - \bm{x}\|^2.
\end{equation}
At this minimizer, the derivative with respect to $t$ must vanish:
\begin{equation}
(\bm{x}_{t^\ast} - \bm{x})^\top \dot{\bm{x}}_{t^\ast} = 0.
\end{equation}
This condition implies that the displacement vector $\bm{x}_{t^\ast} - \bm{x}$, and therefore the gradient $\nabla d(\bm{x})$, is orthogonal to the trajectory tangent $\dot{\bm{x}}_{t^\ast}$:
\begin{equation}
\nabla d(\bm{x})^\top \dot{\bm{x}}_{t^\ast} = 0.
\end{equation}

With this orthogonality property, the Lyapunov derivative reduces to
\begin{equation}
\dot{V}(\bm{x}) = -\lambda_2 d(\bm{x}) \|\nabla d(\bm{x})\|^2 \leq 0,
\end{equation}
with equality only if $d(\bm{x}) = 0$. This shows that the system is globally stable and asymptotically converges to the demonstrated trajectory in the actuated space.

\newpage
\section{GPI algorithm}

\label{app:gpi_algorithm}
\begin{algorithm}[h]
\caption{Geometry-Aware Policy Imitation}
\label{alg:gpi-consistent}
\begin{algorithmic}[1]
\Require $\mathcal{D}=\{\Gamma^{(i)}\}_{i=1}^N$, each $\Gamma^{(i)}=\{(\bm{x}^{(i)}_t,\bm{u}^{(i)}_t)\}_{t=0}^{T_i}$; projection $P$; encoder $\Psi$; robot/environment distances $d_{\mathrm{rob}},d_{\mathrm{env}}$; mixing $\alpha_{\mathrm{rob}},\alpha_{\mathrm{env}}>0$; weights $\lambda_1(\cdot),\lambda_2(\cdot)$; temperature $\beta$; top-$K$
\Ensure Control $\bm{u}\in\mathcal{X}'$ at query $\bm{x}_o$
\State $\bm{x}'_o \gets P(\bm{x}_o)$,\quad $\bm{z}_o \gets \Psi(\bm{x}_o)$
\ForAll{$i\in\{1,\dots,N\}$ \textbf{(parallel over demonstrations)}}
  \State \textbf{Per-time distances}
  \[
    \bm{d}^{(i)}_{\mathrm{rob}} \gets \big(d_{\mathrm{rob}}(\bm{x}'_o,\bm{x}'^{(i)}_t)\big)_{t},\quad
    \bm{d}^{(i)}_{\mathrm{env}} \gets \big(d_{\mathrm{env}}(\bm{z}_o,\Psi(\bm{x}^{(i)}_t))\big)_{t}
  \]
  \State \textbf{Combined distance:}\quad
  $\bm{d}^{(i)} \gets \alpha_{\mathrm{rob}}\bm{d}^{(i)}_{\mathrm{rob}}+\alpha_{\mathrm{env}}\bm{d}^{(i)}_{\mathrm{env}}$
  \State \textbf{Nearest time index and scalar distance:}
  \[
    \kappa^{(i)}(\bm{x}_o) \gets \arg\min_t \bm{d}^{(i)}_t,\qquad
    d(\bm{x}_o\mid\Gamma^{(i)}) \gets \min_t \bm{d}^{(i)}_t
  \]
  \State \textbf{Progression flow:}\;\;
  $\bm{u}^{(i)}_{\kappa} \gets \bm{u}^{(i)}_{\kappa^{(i)}(\bm{x}_o)}=\dot{\bm{x}}'^{(i)}_{\kappa^{(i)}(\bm{x}_o)}$
  \State \textbf{Attraction flow:}\;\;
  $\bm{u}^{(i)}_{\mathrm{att}} \gets -\,\nabla_{\bm{x}'_o}\, d_{\mathrm{rob}}\!\big(\bm{x}'_o,\bm{x}'^{(i)}_{\kappa^{(i)}(\bm{x}_o)}\big)$
  \State \textbf{Local policy:}
  \[
    \pi_i(\bm{x}_o) \;\gets\; \lambda_1\!\big(d(\bm{x}_o\mid\Gamma^{(i)})\big)\,\bm{u}^{(i)}_{\kappa}
    \;+\;
    \lambda_2\!\big(d(\bm{x}_o\mid\Gamma^{(i)})\big)\,\bm{u}^{(i)}_{\mathrm{att}}
  \]
\EndFor
\State \textbf{Top-$K$ selection by demonstration distance:}\;\;
$I_K \gets \text{indices of the $K$ smallest } d(\bm{x}_o\mid\Gamma^{(i)})$
\State \textbf{Softmax weights over selected demos:}\;\;
$w_i(\bm{x}_o) \gets \dfrac{\exp\!\big(-\beta\, d(\bm{x}_o\mid\Gamma^{(i)})\big)}{\sum_{j\in I_K}\exp\!\big(-\beta\, d(\bm{x}_o\mid\Gamma^{(j)})\big)}\;\; (i\in I_K)$
\State \textbf{Global policy:}\;\;
$\displaystyle \bm{u}=\pi(\bm{x}_o)=\sum_{i\in I_K} w_i(\bm{x}_o)\,\pi_i(\bm{x}_o)$
\State \textbf{return } $\bm{u}$
\end{algorithmic}
\label{alg1}
\end{algorithm}

\newpage
\section{Implementation details}

\subsection{PushT task with state-based inputs}  
\label{app:pushT state}
For low-dimensional states, each demonstration is represented as 
\[
\bm{x}_t^{(i)} = [x_a, y_a, x_b, y_b, \theta_b] \in \mathbb{R}^5,
\]  
where $(x_a, y_a)$ denote the agent position, $(x_b, y_b)$ the block position, and $\theta_b$ the block orientation.  
The associated action specifies the target location for a low-level controller:  
\[
\bm{u}_t^{(i)} = [x_{\text{target}}, y_{\text{target}}],
\]  
which we rewrite for velocity control as the relative displacement:  
\[
\bm{u}_t^{(i)} = [x_{\text{target}} - x_a,\; y_{\text{target}} - y_a].
\]  

All state variables are normalized to $[0,1]$ before computing distances.  
The distance field $d(\bm{x}, \Gamma^{(i)})$ is defined as the weighted sum of three components:  
\begin{equation}
d(\bm{x}, \bm{x}_t^{(i)}) 
= w_{\text{obj}} \, \| (x_b, y_b) - (x_b^{(i)}, y_b^{(i)}) \|_2
+ w_{\text{agt}} \, \| (x_a, y_a) - (x_a^{(i)}, y_a^{(i)}) \|_2
+ w_{\theta} \, \mathrm{ang}(\theta_b, \theta_b^{(i)}),
\end{equation}
where $\mathrm{ang}(\cdot, \cdot)$ denotes angular distance.  
Unless otherwise stated, the weights are set to $w_{\text{obj}} = w_{\text{agt}} = w_{\theta} = 1.0$.  

Each demonstration induces a distance field and an associated flow policy.  
At inference time, the global policy is formed by composing the $K$ nearest demonstration policies, with $\lambda_1 = \lambda_2 = 1.0$. Evaluation is performed on environment seeds 500–510 using three distinct policy seeds.

We further explore several variants to improve the flexibility of GPI:

\paragraph{Relative vs. absolute state representation.}  
The PushT task involves nonlinear contact dynamics, so the choice of state representation is important.  
In the \emph{relative} variant, the agent position is expressed in the object’s coordinate frame:  
\begin{equation}
\tilde{\bm{p}}_a = R(-\theta_b)\, \big( (x_a, y_a) - (x_b, y_b) \big),
\end{equation}
where $R(-\theta_b)$ is the $\mathrm{SE}(2)$ rotation matrix aligning the block’s orientation to the $x$-axis.  
The demonstrated action $\bm{u}_t$ is similarly transformed.  
During execution, the predicted action $\tilde{\bm{u}}$ is mapped back to global coordinates via the inverse transformation:  
\begin{equation}
\bm{u} = R(\theta_b) \, \tilde{\bm{u}} + (x_b, y_b).
\end{equation}  

\paragraph{Smooth flow fields.}  
When the action horizon is set to $1$, the controller is highly reactive and may produce abrupt changes whenever the nearest demonstration switches.  
To mitigate this, we apply first-order smoothing to the action sequence:  
\begin{equation}
\bm{u}_t^{\text{smooth}} = \alpha \, \bm{u}_t + (1-\alpha) \, \bm{u}_{t-1}^{\text{smooth}},
\end{equation}
where $\alpha \in [0,1]$ is a smoothing parameter.

\paragraph{Recent-action suppression.}  
To mitigate oscillatory behavior arising from repeatedly selecting near-identical actions,  
we maintain a sliding-window memory $\mathcal{M}$ of the most recent $M$ actions.  
During action selection, if the candidate $\bm{u}_t$ lies within a tolerance $\epsilon$ of any element in $\mathcal{M}$, it is suppressed and the next-best candidate from the composed policy is chosen.  
This mechanism enforces diversity over short horizons, prevents immediate backtracking to previously executed actions,  
and ensures the policy explores novel trajectories while preserving responsiveness.

\paragraph{Perturbed query states.}  
To evaluate robustness, we perturb the query agent position with additive Gaussian noise:  
\begin{equation}
\tilde{\bm{x}}' = \bm{x}' + \epsilon, \qquad \epsilon \sim \mathcal{N}(0, \sigma^2 I),
\end{equation}
where $\bm{x}' = (x_a, y_a)$ is the agent substate.  
The noise variance $\sigma^2$ is annealed over time, decaying from $\sigma=0.1$ at the beginning of execution to $\sigma=0.001$ at later steps.  This perturbation injects stochasticity into the query states, which increases variability in the retrieved flows and can induce multimodal behaviors.

\paragraph{Subsampled demonstrations.}  
For efficiency and robustness, instead of using all demonstrations, we randomly sample a subset $\Gamma_{\text{sub}} \subset \Gamma$ at each query.  
The global policy is then composed over $\Gamma_{\text{sub}}$.  
Empirically, we find that subsampling does not reduce performance; in some cases, the induced stochasticity even helps the agent escape undesirable cycles or “stacked” behaviors.  

\subsection{PushT task with vision-based inputs}  
\label{app:pushT vision}
In the PushT environment, observations consist of an RGB image  
$\mathbf{I}$  
together with agent positions $(x_a, y_a)$.  
Each demonstration state is represented as
\[
\bm{x}_t^{(i)} = [\,x_a, y_a, \mathbf{I}\,].
\]  

\paragraph{Vision encoder.}  
To obtain compact image features, we use an encoder $\psi$ with a ResNet-18 backbone (group normalization) and a projection head (MLP with sizes [512, 256, 128, 3]).  
The encoder is trained with a mean squared error (MSE) loss to predict the object position and orientation:
\[
\psi(\mathbf{I}) \approx [x_o, y_o, \theta_o],
\qquad
\mathcal{L}_{\text{MSE}} = \tfrac{1}{B} \sum_{i=1}^{B} 
\big\|\,\bm{x}_{\text{pred}}^{(i)} - \bm{x}_{\text{target}}^{(i)}\big\|_2^2.
\]
Training is performed for 200 epochs using the Adam optimizer with a learning rate of 0.001.

\paragraph{Distance metric and policy synthesis.}  
After training, each demonstration image is embedded as
\[
\bm{z}_t^{(i)} = \psi(\mathbf{I}_t^{(i)}),
\]
and for a query state $\bm{x}_o = [x_a, y_a, \mathbf{I}]$,
\[
\bm{z}_o = \psi(\mathbf{I}).
\]  
Distances are defined in this learned feature space and 
policy synthesis then proceeds identically to the state-based inputs.

\subsection{PushT task with ResNet-18 encoder and PCA}
\label{app:resnet_pca}
We construct a compact observation embedding by reusing the same ResNet-18 encoder from the Diffusion Policy implementation (task-pretrained on \textsc{PushT}). At inference, this encoder is frozen and used as a fixed feature extractor. We aggregate features over a short temporal window (\texttt{obs\_horizon} $=2$), apply PCA for dimensionality reduction on the image features, and concatenate with the last two agent positions (normalized and reweighted to balance scale). Each demonstration is thus represented in this joint embedding space. At test time, the current observation is embedded in the same way, and the closest demonstration under cosine similarity is identified. The policy then follows the flow induced by this demonstration, with progression and attraction weights set to $\lambda_1 = \lambda_2 = 1.0$.

\paragraph{Per-timestep features.}
Given an image $\bm{I} $ and agent position $[x_a,y_a]$, we extract a 512-D descriptor $\psi(\bm{I})$ with the frozen ResNet-18 backbone (final FC removed; BatchNorm $\rightarrow$ GroupNorm as in the diffusion policy). 

\paragraph{Temporal windowing and dimensionality reduction.}
With \texttt{obs\_horizon} $T=2$, we flatten the last $T$ descriptors and apply IncrementalPCA to project them to $16$ principal components:
\[
\bm{z}_t \;=\; \mathrm{PCA}_{16}([\psi(I_{t-1}), \psi(I_t)]) \in \mathbb{R}^{16}.
\]

\paragraph{Concatenation with agent positions.}
To balance image and agent information, we concatenate the PCA embedding $z_t$ with the normalized agent positions from the last two steps. All embeddings are L2-normalized before similarity computations.

\paragraph{Policy selection.}
At test time, the query embedding is compared to the demonstration database using cosine similarity, and the flow is executed with $\lambda_1=\lambda_2=1.0$. To prevent degenerate repeats, the selected pair is removed from the database at the next step.

\subsection{PushT task with VAE}
\label{app:vae}
We construct a compact observation embedding using a convolutional variational autoencoder (VAE) trained directly on \textsc{PushT} images. At inference, we discard the decoder and use only the encoder to produce latent codes, which are concatenated with scaled agent positions to form the final embedding. The global policy then follows the flow induced by the closest demonstration under cosine similarity, with progression and attraction weights set to $\lambda_1 = \lambda_2 = 1.0$.

\paragraph{Per-timestep features.}
Given an image $\bm{I}_t$ with pixel values normalized to $[0,1]$, the VAE encoder outputs a Gaussian posterior
\[
\bm{z}_t \;\sim\; q_\phi(\bm{z} \mid \bm{I}_t), \quad \bm{z}_t \in \mathbb{R}^d,
\]
with diagonal covariance. At inference, we use only the posterior mean $\mu_t$ as the latent feature. 


\paragraph{Retrieval.}
At test time, we encode the current observation window to obtain $\bm{z}_t$, normalize it, and compute cosine similarity against the stored database features. The demonstration with the highest similarity is selected, and its associated action sequence defines the flow. Cosine similarity achieved slightly higher performance (average return $\approx 0.88$) compared to Euclidean distance ($\approx 0.85$).

\paragraph{Training Setup.}
We train the VAE with a standard Gaussian prior \(p(\mathbf{z})=\mathcal{N}(\mathbf{0}, I)\) and a Gaussian reconstruction likelihood \(p(\mathbf{x}\mid \mathbf{z})=\mathcal{N}\!\big(\hat{\mathbf{x}}(\mathbf{z}),\, \tau^{2} I\big)\) with fixed \(\tau = 2 \times 10^{-1}\).
This choice of \(\tau\) balanced the reconstruction and KL terms: with \(\tau = 0.2\) both the reconstruction loss and the KL divergence decreased steadily, whereas using smaller \(\tau\) values led to optimization stalling (neither term decreased).
Training was performed for 25 epochs with the Adam optimizer (learning rate \(1\times 10^{-4}\)).
At inference, we discard the decoder and use only the encoder’s posterior mean.

\subsection{PushT task with SAM-based pose embedding}
\label{app:sam}

We estimate object pose directly from images using a pretrained SAM/SAM2 pipeline (no fine-tuning). From each frame we obtain a binary mask of the T-block, from which we extract its centroid \((x_b, y_b)\) and axial orientation \(\theta_b\) (defined modulo \(\pi\)). Combined with the agent position \((x_a, y_a)\), this yields the state
\[
\bm{x}_t = [\,x_a, y_a, x_b, y_b, \theta_b\,] \in \mathbb{R}^5.
\]

All variables are normalized to \([0,1]\) before distance computations; angular differences use the same axial angular distance as in the state-based setup. Distances and policy composition follow the same formulation, with weights \(w_{\text{obj}}=w_{\text{agt}}=w_{\theta}=1.0\) and flow execution with \(\lambda_1=\lambda_2=1.0\).

\paragraph{Per-timestep pose extraction.}
Given a SAM mask, the centroid is
\[
(x_b, y_b) \;=\; \mathrm{centroid}(\text{mask}),
\]
and the orientation is computed from second-order moments of foreground pixels. Let \(\mu_{pq}\) denote centralized moments; the principal axis corresponding to the largest covariance eigenvalue indicates the elongation direction. We define
\[
\theta_b \;=\; \tfrac{1}{2}\operatorname{atan2}\!\big(2\mu_{11},\, \mu_{20}-\mu_{02}+\varepsilon\big),
\]
wrap \(\theta_b\) to \((-\pi,\pi]\), and treat it as axial (modulo \(\pi\)) for angular distance.

\paragraph{Retrieval and policy selection.}
At test time, we form \(\bm{x}_t=[x_a,y_a,x_b,y_b,\theta_b]\), apply the same normalization as above, and compute distances to all stored demonstration states using the state-based metric. We retrieve the \(K\) nearest neighbors (default \(K=1\)) and execute the composed flow with \(\lambda_1=\lambda_2=1.0\).

\paragraph{Tracking and prompting details.}
We use SAM2’s video predictor (\texttt{sam2.1\_hiera\_tiny}) to track the T-block across frames, re-prompting each step with a skeletal outline derived from the most recent pose estimate to stabilize mask propagation. To compensate for a small systematic bias in predicted centroids, we apply a constant offset correction to \((x_b,y_b)\), calibrated on seeds 500--700.

\paragraph{Limitations.}
Performance depends on segmentation quality; occlusions and viewpoint changes can induce drift in the estimated pose, which in turn affects retrieval and control.

\subsection{Robot-flip task}
\label{app:robot-flip}
\textbf{Robot teleoperation}: We utilized a bimanual robotic system configured with a ViperX300s (follower) and a WidowX250 (leader), along with a RealSense D405 camera from a top-down view. The system is built on an open-source platform. By using robot teleoperation, we collected 121 demonstrations, each contains 200 to 1000 timesteps to complete the flip task. The dataset is structured in an HDF5 format and includes robot actions and observations, where observations are composed of effort, images, joint angles, and joint velocities. Specifically, we teleoperated the leader robot (WidowX250) to control the follower (ViperX300s) robot for manipulation tasks (flip the box). The camera records images at an 848\(\times\)480 resolution with a 30 Hz frequency, and then crops them to a 320\(\times\)240 resolution for policy training. 

\textbf{Policy imitation.} The policy imitation process is similar to the pushT task with vision-based inputs. Specifically, we use a vision encoder that takes RGB images as input and predicts the desired robot action as a latent embedding using an MSE loss. Training is performed for 100 epochs using the Adam optimizer with a learning rate of 0.0001. After training, we calculate the latent feature of each demonstrated image as a feature database. The online inference involves the computation of a distance field that includes both distance measurement in this latent space and an additional distance metric for joint position displacement, guiding the flow field and policy composition. Both attraction and progression parameters are set to 1.0 during execution. To ensure the temporal consistency, the task is run with horizon=100.

\begin{figure}[htbp]\centering
    \includegraphics[width=0.5\textwidth]{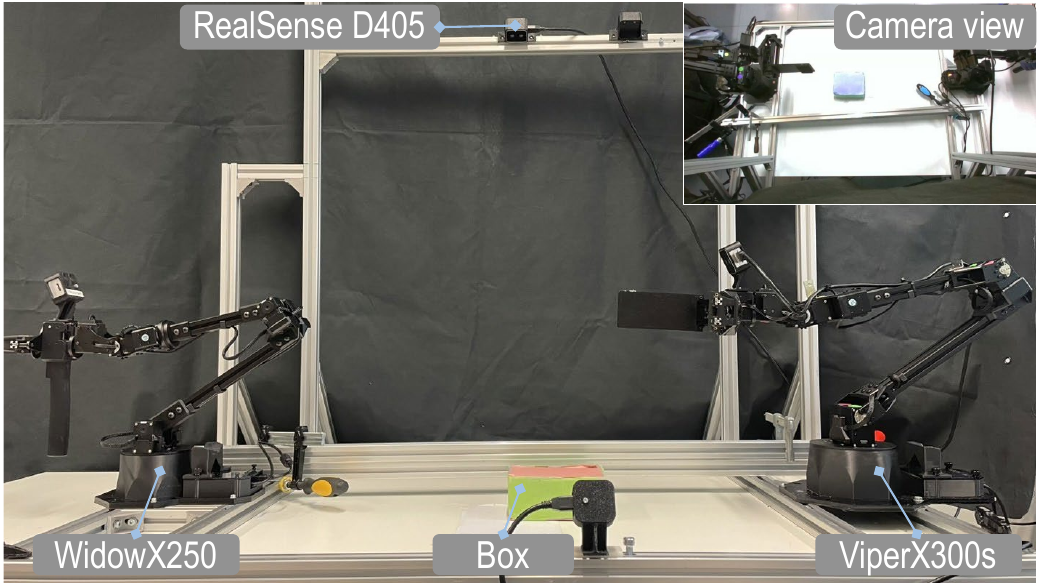}
    \caption{ALOHA teleoperation platform.}
    \label{fig:robot_platform}   
\end{figure} 

\subsection{Human–robot interaction task}
\label{app:hri}
We use the \texttt{openai/clip-vit-base-patch32} CLIP model for vision–language grounding.
Positive and negative text prompts for hand–held object detection are listed below.

\paragraph{Text prompts.}
\begin{lstlisting}[language=Python, basicstyle=\ttfamily\small]
pos_prompts = [
    "a photo of a hand holding a banana",
    "a hand holding an apple",
    "a human hand holding an orange",
    "a hand holding a pear",
    "a hand holding a strawberry",
    "a hand holding grapes",
    "a hand holding a piece of fruit",
    "a person's hand holding a fruit",
    "close-up of a hand holding a fruit",
]

neg_prompts = [
    "an empty hand",
    "a hand with nothing in it",
    "a hand holding a baseball",
    "a hand holding a black ball",
    "a hand holding a blue cup",
    "a hand holding a plastic cup",
    "a hand holding adhesive tape",
    "a hand holding a tape roll",
    "a hand holding a screwdriver",
    "a hand holding a tool",
    "a hand holding a non-fruit object",
]
\end{lstlisting}

\newpage
\section{Additional experimental results}

\subsection{Memory cost}
\label{app:memory_cost}

The state-based \textsc{PushT} dataset has $25{,}000 \times 7 = 175{,}000$ elements, requiring $175{,}000 \times 4 \approx 0.67$ MB with \texttt{float32}, consistent with the observed 0.7 MB. For comparison, an MLP with layers $[7,512,256,128,1]$ has $168{,}449$ parameters ($\approx 0.64$ MB), which is at a similar scale. However, typical models are far larger than simple MLP; e.g., a state-based diffusion policy exceeds $200$ MB.

Although GPI’s memory grows linearly with the number of demonstrations, this is practical in our setting: robot actions are low-dimensional, and high-dimensional observations are stored as compact latent features. Inference is lightweight, parallelizable, and can use subsampling or approximate nearest-neighbor search to bound latency. 
As we demonstrated in the paper, GPI achieves orders-of-magnitude gains in efficiency over standard baselines in common imitation-learning settings.

\subsection{Robomimic and Adroit Hand tasks}
\label{app:other_tasks}

\begin{figure}[h]
    \centering
    \includegraphics[width=0.9\textwidth]{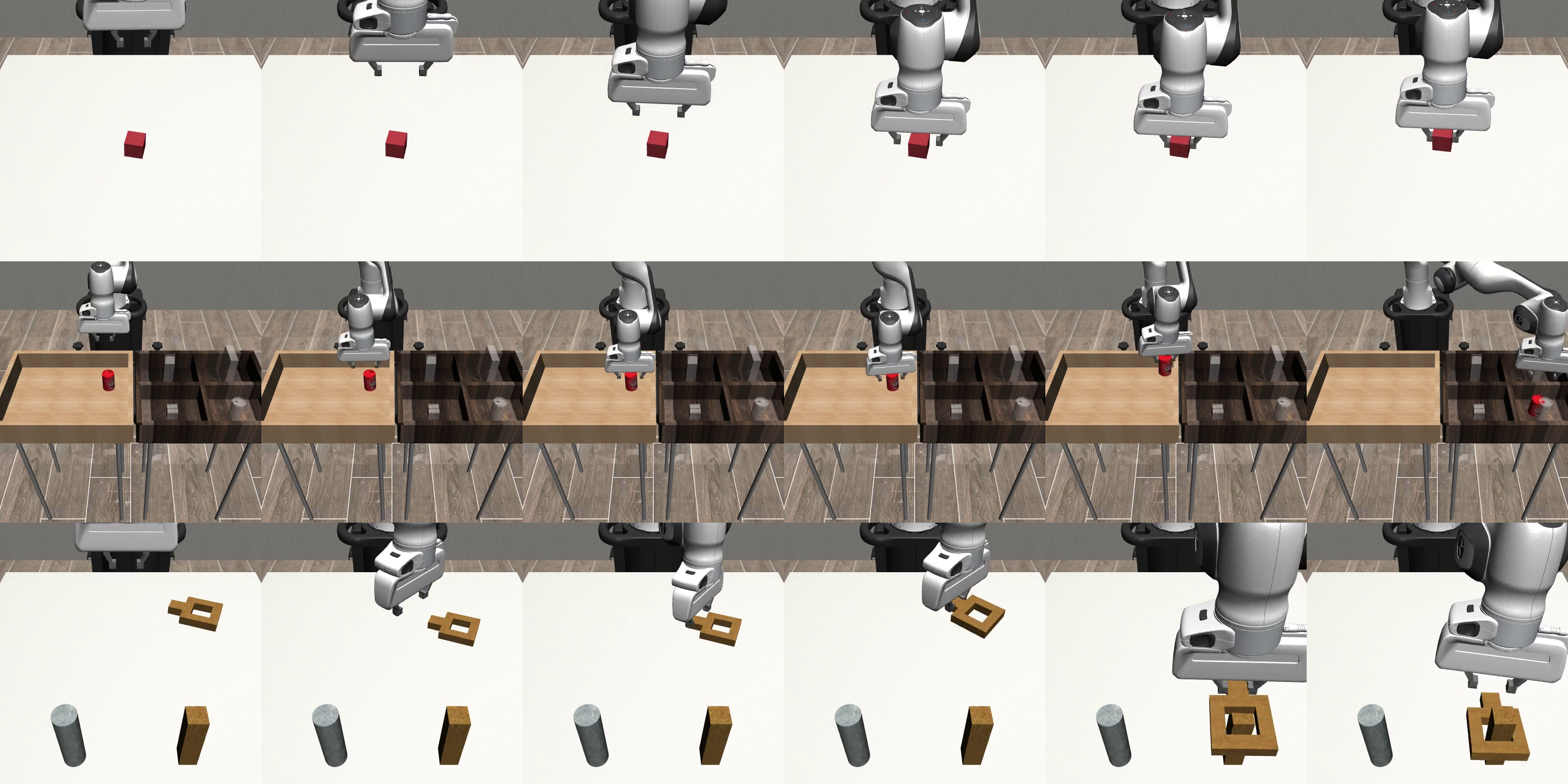}
    \caption{Snapshots of experimental results for Lift, Can, and Square tasks on Robomimic environments.}
    \label{fig:robomimic_exp}    
\end{figure} 

\begin{figure}[h]
    \centering
    \includegraphics[width=0.9\textwidth]{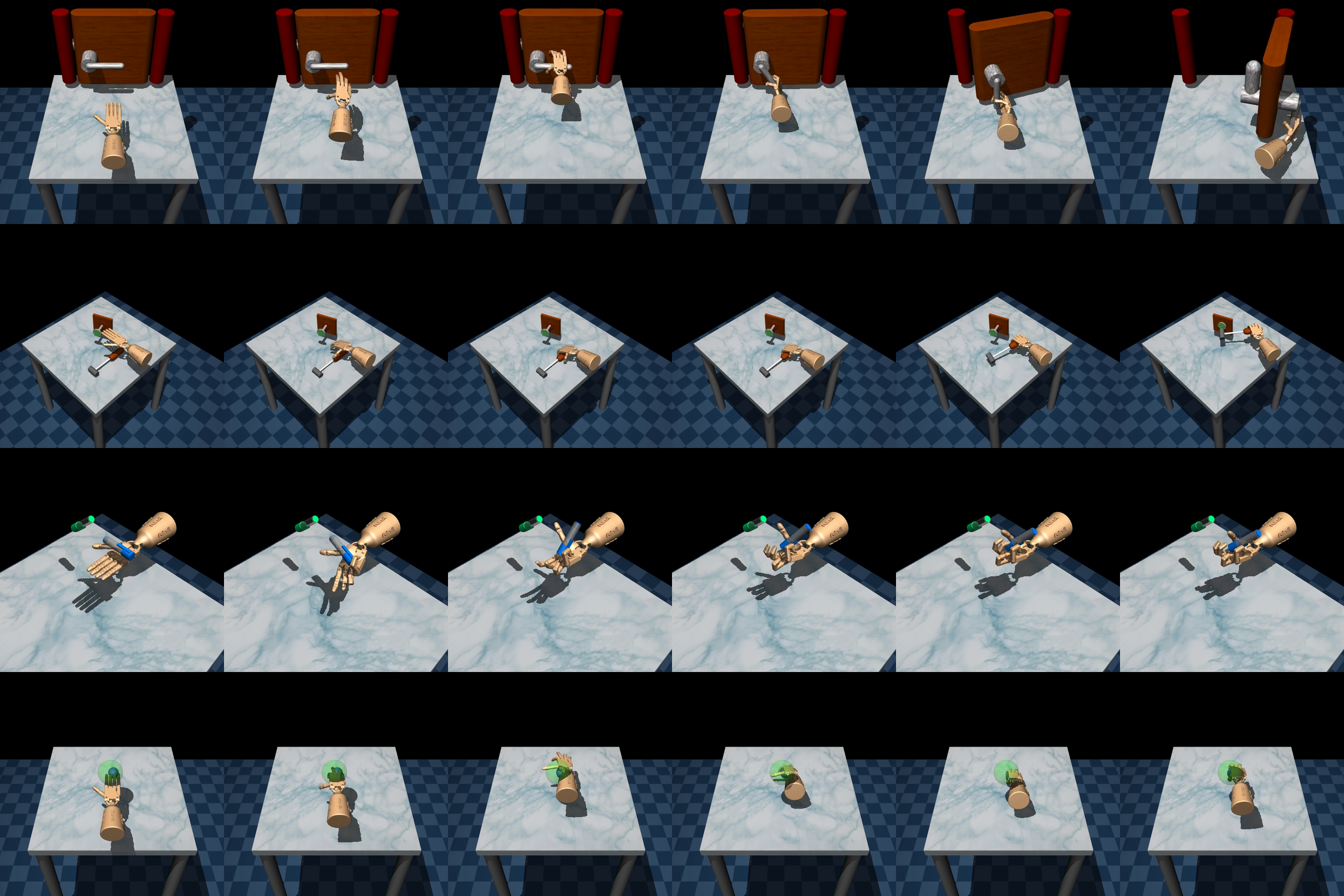}
    \caption{Snapshots of experimental results for Door, Hammer, Pen, and Relocate on Adroit hand tasks.}
    \label{fig:adroit_hand_task}    
\end{figure} 


\newpage
\subsection{2D maze}
\label{app:2d_maze}
We evaluate our approach on the 2D Maze benchmark, previously used by \citep{chen2025diffusion, janner2022diffuser}. Unlike these methods, our approach is \emph{training-free}: at test time we select a suffix of a single demonstration using a simple distance metric and execute it. Concretely, for demonstration \(i\) of length \(H\) and timestep \(k\), we minimize
\begin{equation*}
D(i,k)\;=\;10\,\lVert \mathbf{x}_0-\mathbf{x}^{(i)}_{k}\rVert_2 \;+\; 5\,\lVert \mathbf{x}_g-\mathbf{x}^{(i)}_{g}\rVert_2 \;+\; 0.1\,(H-k),
\label{eq:maze_distance}
\end{equation*}
where \(\mathbf{x}_0\) is the initial state, \(\mathbf{x}^{(i)}_{k}\) is the \(k\)-th state of demonstration \(i\), \(\mathbf{x}_g\) is the task goal, and \(\mathbf{x}^{(i)}_{g}\) is the goal state associated with demonstration \(i\). The final term penalizes long remaining horizons; since 2D Maze demonstrations can include detours, this bias favors suffixes that proceed more directly to the goal. After selecting \((i^\star,k^\star)\), we execute the suffix \(\{\mathbf{x}^{(i^\star)}_{k^\star:H}\}\) as the plan. In doing so, our method also recovers the effective task horizon \(H - k^\star\), something most alternative approaches cannot determine directly. Instead, they must either: (i) assume a long horizon and truncate once the task is completed, (ii) assume a short horizon and repeat until completion, or (iii) try multiple horizons and select the smallest successful one.

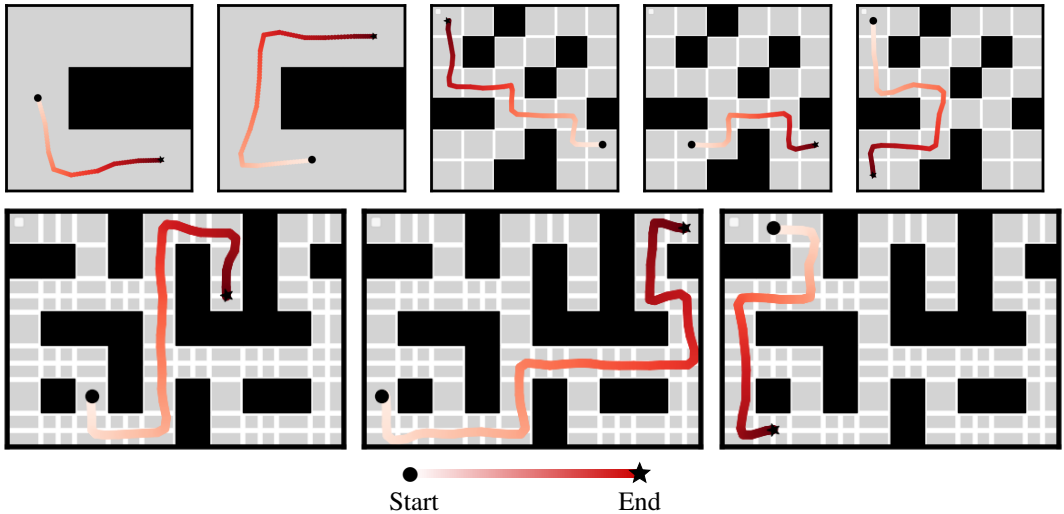
\begin{figure}[h]
\centering

\begin{minipage}{0.185\linewidth}
    \includegraphics[width=\linewidth, angle=90]{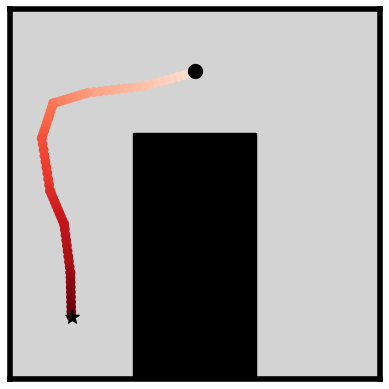}
\end{minipage}
\hspace{0.005\linewidth}
\begin{minipage}{0.185\linewidth}
    \includegraphics[width=\linewidth, angle=90]{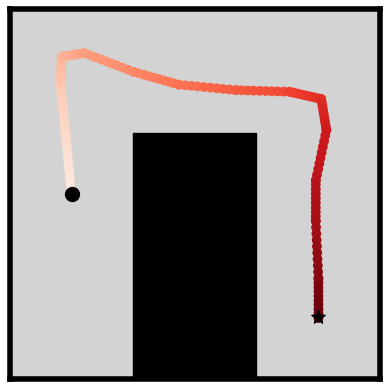}
\end{minipage}
\hspace{0.005\linewidth}
\begin{minipage}{0.185\linewidth}
    \includegraphics[width=\linewidth, angle=90]{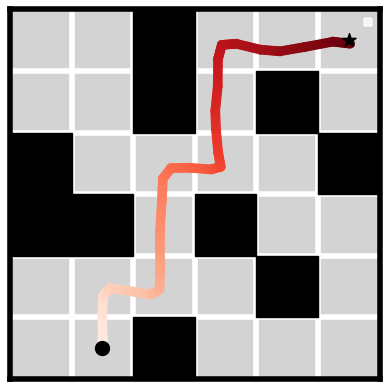}
\end{minipage}
\hspace{0.005\linewidth}
\begin{minipage}{0.185\linewidth}
    \includegraphics[width=\linewidth, angle=90]{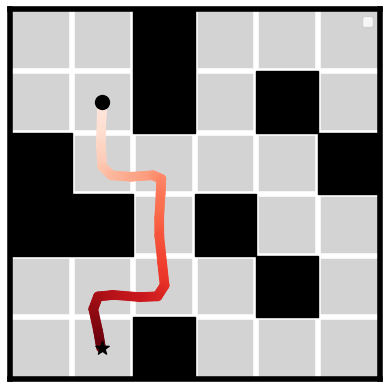}
\end{minipage}
\hspace{0.005\linewidth}
\begin{minipage}{0.185\linewidth}
    \includegraphics[width=\linewidth, angle=90]{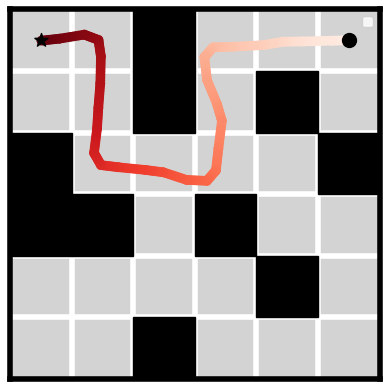}
\end{minipage}

\vspace{0.2em} 

\begin{minipage}{0.32\linewidth}
    \includegraphics[width=0.75\linewidth, angle=90]{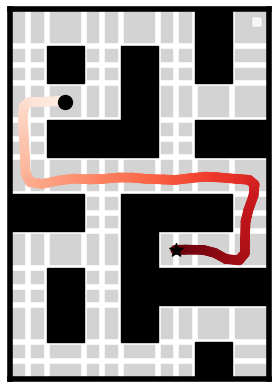}
\end{minipage}
\hfill
\begin{minipage}{0.32\linewidth}
    \includegraphics[width=0.75\linewidth, angle = 90]{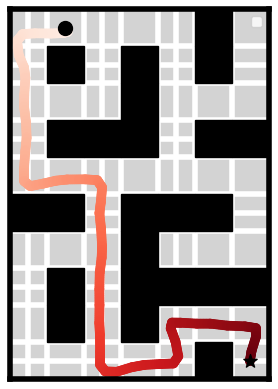}
\end{minipage}
\hfill
\begin{minipage}{0.32\linewidth}
    \includegraphics[width=0.75\linewidth, angle = 90]{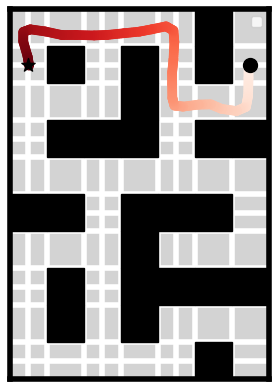}
\end{minipage}

\begin{tikzpicture}
  \shade[left color=white,right color=red!80!black]
    (0,0.075) rectangle (3,0.175);

  \node[anchor=east] at (0.2,0.1) {\textcolor{black}{\Large $\bullet$}};
  \node[anchor=north] at (0,0) {Start};

  \node[
  star, star points=5, star point ratio=2.25,
  minimum size=8pt, inner sep=0pt,
  draw=black, fill=black,
  anchor=south
] at (3,0.05) {};
  \node[anchor=north] at (3,0) {End};
\end{tikzpicture}

\caption{Results on 2D Maze using our method. Without any training, a simple distance-based criterion achieves a 100\% success rate across all tasks, with an average inference time of 0.08 seconds.}
\label{fig:2d_maze_res}
\end{figure}

\newpage
\section{Reproducibility Statement}
We will release our code, configuration files, and evaluation scripts upon publication. Key implementation details and protocols are documented in the main text and appendix to facilitate reproduction in the interim.

\section{Use of Large Language Models (LLMs)}
We used LLMs (e.g., ChatGPT and Claude) to rephrase and polish the manuscript and to assist with coding tasks. All LLM-generated code was reviewed, edited, and integrated by the authors; the LLM did not design algorithms or produce experimental results.

%% file: main.bbl
\begin{thebibliography}{35}
\providecommand{\natexlab}[1]{#1}
\providecommand{\url}[1]{\texttt{#1}}
\expandafter\ifx\csname urlstyle\endcsname\relax
  \providecommand{\doi}[1]{doi: #1}\else
  \providecommand{\doi}{doi: \begingroup \urlstyle{rm}\Url}\fi

\bibitem[Aldaco et~al.(2024)Aldaco, Armstrong, Baruch, Bingham, Chan, Draper, Dwibedi, Finn, Florence, Goodrich, et~al.]{aldaco2024aloha}
Jorge Aldaco, Travis Armstrong, Robert Baruch, Jeff Bingham, Sanky Chan, Kenneth Draper, Debidatta Dwibedi, Chelsea Finn, Pete Florence, Spencer Goodrich, et~al.
\newblock Aloha 2: An enhanced low-cost hardware for bimanual teleoperation.
\newblock \emph{arXiv preprint arXiv:2405.02292}, 2024.

\bibitem[Barreiros et~al.(2025)Barreiros, Beaulieu, Bhat, Cory, Cousineau, Dai, Fang, Hashimoto, Irshad, Itkina, et~al.]{barreiros2025careful}
Jose Barreiros, Andrew Beaulieu, Aditya Bhat, Rick Cory, Eric Cousineau, Hongkai Dai, Ching-Hsin Fang, Kunimatsu Hashimoto, Muhammad~Zubair Irshad, Masha Itkina, et~al.
\newblock A careful examination of large behavior models for multitask dexterous manipulation.
\newblock \emph{arXiv preprint arXiv:2507.05331}, 2025.

\bibitem[Bertsekas(1995)]{bertsekas1995dynamic}
Dimitri~P. Bertsekas.
\newblock \emph{Dynamic Programming and Optimal Control, Volumes I and II}.
\newblock Athena Scientific, Belmont, MA, 1st edition, 1995.

\bibitem[Calinon et~al.(2007)Calinon, Guenter, and Billard]{calinon2007learning}
Sylvain Calinon, Florent Guenter, and Aude Billard.
\newblock On learning, representing, and generalizing a task in a humanoid robot.
\newblock \emph{IEEE Transactions on Systems, Man, and Cybernetics, Part B (Cybernetics)}, 37\penalty0 (2):\penalty0 286--298, 2007.

\bibitem[Chen et~al.(2025)Chen, Mart{\'\i}~Mons{\'o}, Du, Simchowitz, Tedrake, and Sitzmann]{chen2025diffusion}
Boyuan Chen, Diego Mart{\'\i}~Mons{\'o}, Yilun Du, Max Simchowitz, Russ Tedrake, and Vincent Sitzmann.
\newblock Diffusion forcing: Next-token prediction meets full-sequence diffusion.
\newblock \emph{Advances in Neural Information Processing Systems}, 37:\penalty0 24081--24125, 2025.

\bibitem[Chi et~al.(2023)Chi, Xu, Feng, Cousineau, Du, Burchfiel, Tedrake, and Song]{chi2023diffusion}
Cheng Chi, Zhenjia Xu, Siyuan Feng, Eric Cousineau, Yilun Du, Benjamin Burchfiel, Russ Tedrake, and Shuran Song.
\newblock Diffusion policy: Visuomotor policy learning via action diffusion.
\newblock \emph{The International Journal of Robotics Research}, pp.\  02783649241273668, 2023.

\bibitem[Du \& Mordatch(2019)Du and Mordatch]{du2019implicit}
Yilun Du and Igor Mordatch.
\newblock Implicit generation and generalization in energy-based models.
\newblock In \emph{Advances in Neural Information Processing Systems (NeurIPS)}, volume~32, 2019.

\bibitem[Florence et~al.(2022)Florence, Lynch, Zeng, Ramirez, Wahid, Downs, Wong, Lee, Mordatch, and Tompson]{florence2022implicit}
Pete Florence, Corey Lynch, Andy Zeng, Oscar~A Ramirez, Ayzaan Wahid, Laura Downs, Adrian Wong, Johnny Lee, Igor Mordatch, and Jonathan Tompson.
\newblock Implicit behavioral cloning.
\newblock In \emph{Conference on robot learning}, pp.\  158--168. PMLR, 2022.

\bibitem[He et~al.(2016)He, Zhang, Ren, and Sun]{he2016deep}
Kaiming He, Xiangyu Zhang, Shaoqing Ren, and Jian Sun.
\newblock Deep residual learning for image recognition.
\newblock \emph{Proceedings of the IEEE conference on computer vision and pattern recognition}, pp.\  770--778, 2016.

\bibitem[Ho et~al.(2020)Ho, Jain, and Abbeel]{ho2020denoising}
Jonathan Ho, Ajay Jain, and Pieter Abbeel.
\newblock Denoising diffusion probabilistic models.
\newblock \emph{Advances in neural information processing systems}, 33:\penalty0 6840--6851, 2020.

\bibitem[Hotelling(1933)]{hotelling1933pca}
Harold Hotelling.
\newblock Analysis of a complex of statistical variables into principal components.
\newblock \emph{Journal of Educational Psychology}, 24\penalty0 (6):\penalty0 417--441, 1933.

\bibitem[Ijspeert et~al.(2013)Ijspeert, Nakanishi, Hoffmann, Pastor, and Schaal]{ijspeert2013dynamical}
Auke~Jan Ijspeert, Jun Nakanishi, Heiko Hoffmann, Peter Pastor, and Stefan Schaal.
\newblock Dynamical movement primitives: learning attractor models for motor behaviors.
\newblock \emph{Neural computation}, 25\penalty0 (2):\penalty0 328--373, 2013.

\bibitem[Intelligence et~al.(2025)Intelligence, Black, Brown, Darpinian, Dhabalia, Driess, Esmail, Equi, Finn, Fusai, et~al.]{intelligence2025pi}
Physical Intelligence, Kevin Black, Noah Brown, James Darpinian, Karan Dhabalia, Danny Driess, Adnan Esmail, Michael Equi, Chelsea Finn, Niccolo Fusai, et~al.
\newblock \(\ \backslash$pi\_ $\{$0.5$\}\)\: a vision-language-action model with open-world generalization.
\newblock \emph{arXiv preprint arXiv:2504.16054}, 2025.

\bibitem[Janner et~al.(2022)Janner, Du, Tenenbaum, and Levine]{janner2022diffuser}
Michael Janner, Yilun Du, Joshua Tenenbaum, and Sergey Levine.
\newblock Planning with diffusion for flexible behavior synthesis.
\newblock In \emph{International Conference on Machine Learning}, 2022.

\bibitem[Jiang et~al.(2025)Jiang, Fang, Roy, Lozano-P{\'e}rez, Kaelbling, and Ancha]{jiang2025streaming}
Sunshine Jiang, Xiaolin Fang, Nicholas Roy, Tom{\'a}s Lozano-P{\'e}rez, Leslie~Pack Kaelbling, and Siddharth Ancha.
\newblock Streaming flow policy: Simplifying diffusion $/$ flow-matching policies by treating action trajectories as flow trajectories.
\newblock \emph{arXiv preprint arXiv:2505.21851}, 2025.

\bibitem[Khansari-Zadeh \& Billard(2011)Khansari-Zadeh and Billard]{khansari2011learning}
S~Mohammad Khansari-Zadeh and Aude Billard.
\newblock Learning stable nonlinear dynamical systems with {G}aussian mixture models.
\newblock \emph{IEEE Transactions on Robotics}, 27\penalty0 (5):\penalty0 943--957, 2011.

\bibitem[Kingma \& Welling(2013)Kingma and Welling]{kingma2013auto}
Diederik~P Kingma and Max Welling.
\newblock Auto‐encoding variational bayes.
\newblock \emph{arXiv preprint arXiv:1312.6114}, 2013.

\bibitem[Kirillov et~al.(2023)Kirillov, Mintun, Ravi, Mao, Rolland, Gustafson, Xiao, Whitehead, Berg, Lo, et~al.]{kirillov2023segment}
Alexander Kirillov, Eric Mintun, Nikhila Ravi, Hanzi Mao, Chloe Rolland, Laura Gustafson, Tete Xiao, Spencer Whitehead, Alexander~C Berg, Wan-Yen Lo, et~al.
\newblock Segment anything.
\newblock In \emph{2023 IEEE/CVF International Conference on Computer Vision (ICCV)}, pp.\  3992--4003. IEEE Computer Society, 2023.

\bibitem[LeCun et~al.(2006)LeCun, Chopra, Hadsell, Ranzato, Huang, et~al.]{lecun2006tutorial}
Yann LeCun, Sumit Chopra, Raia Hadsell, M~Ranzato, Fujie Huang, et~al.
\newblock A tutorial on energy-based learning.
\newblock \emph{Predicting structured data}, 1\penalty0 (0), 2006.

\bibitem[Li \& Calinon(2025)Li and Calinon]{Li25MPDS}
Y.~Li and S.~Calinon.
\newblock From movement primitives to distance fields to dynamical systems.
\newblock \emph{{IEEE} Robotics and Automation Letters ({RA-L})}, 2025.

\bibitem[Lipman et~al.(2023)Lipman, Chen, Ben-Hamu, Nickel, and Le]{lipmanflow}
Yaron Lipman, Ricky~TQ Chen, Heli Ben-Hamu, Maximilian Nickel, and Matt Le.
\newblock Flow matching for generative modeling.
\newblock In \emph{11th International Conference on Learning Representations, ICLR 2023}, 2023.

\bibitem[Mandlekar et~al.(2021)Mandlekar, Xu, Wong, Nasiriany, Wang, Kulkarni, Fei-Fei, Savarese, Zhu, and Mart\'{i}n-Mart\'{i}n]{robomimic2021}
Ajay Mandlekar, Danfei Xu, Josiah Wong, Soroush Nasiriany, Chen Wang, Rohun Kulkarni, Li~Fei-Fei, Silvio Savarese, Yuke Zhu, and Roberto Mart\'{i}n-Mart\'{i}n.
\newblock What matters in learning from offline human demonstrations for robot manipulation.
\newblock In \emph{Conference on Robot Learning (CoRL)}, 2021.

\bibitem[Nair et~al.(2018)Nair, McGrew, Andrychowicz, Zaremba, and Abbeel]{Nair2017OvercomingEI}
Ashvin Nair, Bob McGrew, Marcin Andrychowicz, Wojciech Zaremba, and P.~Abbeel.
\newblock Overcoming exploration in reinforcement learning with demonstrations.
\newblock \emph{International Conference on Robotics and Automation (ICRA)}, pp.\  6292--6299, 2018.

\bibitem[Nijkamp et~al.(2020)Nijkamp, Hill, Han, Zhu, and Wu]{nijkamp2020ebm}
Erik Nijkamp, Mitch Hill, Tian Han, Song-Chun Zhu, and Ying~Nian Wu.
\newblock Learning non-convergent non-persistent short-run {MCMC} toward energy-based model.
\newblock In \emph{Advances in Neural Information Processing Systems (NeurIPS)}, volume~33, pp.\  11588--11600, 2020.

\bibitem[Osa et~al.(2018)Osa, Pardo, Neumann, Bagnell, Abbeel, and Peters]{osa2018algorithmic}
Takayuki Osa, Fabio Pardo, Gerhard Neumann, J.~Andrew Bagnell, Pieter Abbeel, and Jan Peters.
\newblock An algorithmic perspective on imitation learning.
\newblock \emph{Foundations and Trends in Robotics}, 7\penalty0 (1-2):\penalty0 1--179, 2018.
\newblock \doi{10.1561/2300000053}.

\bibitem[Paraschos et~al.(2013)Paraschos, Daniel, Peters, and Neumann]{paraschos2013probabilistic}
Alexandros Paraschos, Christian Daniel, Jan~R Peters, and Gerhard Neumann.
\newblock Probabilistic movement primitives.
\newblock \emph{Advances in neural information processing systems}, 26, 2013.

\bibitem[Pari et~al.(2022)Pari, Shafiullah, Arunachalam, and Pinto]{Pari-RSS-22}
Jyothish Pari, {Nur Muhammad (Mahi)} Shafiullah, {Sridhar Pandian} Arunachalam, and Lerrel Pinto.
\newblock {The Surprising Effectiveness of Representation Learning for Visual Imitation}.
\newblock In \emph{Proceedings of Robotics: Science and Systems}, New York City, NY, USA, June 2022.
\newblock \doi{10.15607/RSS.2022.XVIII.010}.

\bibitem[Radford et~al.(2021)Radford, Kim, Hallacy, Ramesh, Goh, Agarwal, Sastry, Askell, Mishkin, Clark, Krueger, and Sutskever]{radford2021learning}
Alec Radford, Jong~Wook Kim, Chris Hallacy, Aditya Ramesh, Gabriel Goh, Sandhini Agarwal, Girish Sastry, Amanda Askell, Pamela Mishkin, Jack Clark, Gretchen Krueger, and Ilya Sutskever.
\newblock Learning transferable visual models from natural language supervision.
\newblock In \emph{Proceedings of the 38th International Conference on Machine Learning}, volume 139 of \emph{ICML}, pp.\  8748--8763, 2021.

\bibitem[Rajeswaran et~al.(2018)Rajeswaran, Kumar, Gupta, Vezzani, Schulman, Todorov, and Levine]{rajeswaran2018learning}
Aravind Rajeswaran, Vikash Kumar, Abhishek Gupta, Giulia Vezzani, John Schulman, Emanuel Todorov, and Sergey Levine.
\newblock Learning complex dexterous manipulation with deep reinforcement learning and demonstrations.
\newblock \emph{Robotics: Science and Systems XIV}, 2018.

\bibitem[Razmjoo et~al.(2021)Razmjoo, Lembono, and Calinon]{razmjoo2021optimal}
A.~Razmjoo, T.~S. Lembono, and S.~Calinon.
\newblock Optimal control combining emulation and imitation to acquire physical assistance skills.
\newblock In \emph{20th International Conference on Advanced Robotics (ICAR)}, pp.\  338--343. IEEE, 2021.

\bibitem[Sim{\'e}oni et~al.(2025)Sim{\'e}oni, Vo, Seitzer, Baldassarre, Oquab, Jose, Khalidov, Szafraniec, Yi, Ramamonjisoa, Massa, Haziza, Wehrstedt, Wang, Darcet, Moutakanni, Sentana, Roberts, Vedaldi, Tolan, Brandt, Couprie, Mairal, J{\'e}gou, Labatut, and Bojanowski]{simeoni2025dinov3}
Oriane Sim{\'e}oni, Huy~V. Vo, Maximilian Seitzer, Federico Baldassarre, Maxime Oquab, Cijo Jose, Vasil Khalidov, Marc Szafraniec, Seungeun Yi, Micha{\"e}l Ramamonjisoa, Francisco Massa, Daniel Haziza, Luca Wehrstedt, Jianyuan Wang, Timoth{\'e}e Darcet, Th{\'e}o Moutakanni, Leonel Sentana, Claire Roberts, Andrea Vedaldi, Jamie Tolan, John Brandt, Camille Couprie, Julien Mairal, Herv{\'e} J{\'e}gou, Patrick Labatut, and Piotr Bojanowski.
\newblock {DINOv3}, 2025.
\newblock URL \url{https://arxiv.org/abs/2508.10104}.

\bibitem[Song et~al.(2021)Song, Meng, and Ermon]{song2020denoising}
Jiaming Song, Chenlin Meng, and Stefano Ermon.
\newblock Denoising diffusion implicit models.
\newblock In \emph{International Conference on Learning Representations}, 2021.

\bibitem[Song \& Ermon(2019)Song and Ermon]{song2019generative}
Yang Song and Stefano Ermon.
\newblock Generative modeling by estimating gradients of the data distribution.
\newblock In \emph{Advances in Neural Information Processing Systems (NeurIPS)}, volume~32, 2019.

\bibitem[Sutton \& Barto(1998)Sutton and Barto]{sutton1998reinforcement}
Richard~S. Sutton and Andrew~G. Barto.
\newblock \emph{Reinforcement Learning: An Introduction}.
\newblock MIT Press, Cambridge, MA, 1st edition, 1998.

\bibitem[Zhang \& Gienger(2024)Zhang and Gienger]{zhang2024affordance}
Fan Zhang and Michael Gienger.
\newblock Affordance-based robot manipulation with flow matching.
\newblock \emph{arXiv preprint arXiv:2409.01083}, 2024.

\end{thebibliography}
